\documentclass{article}

\PassOptionsToPackage{numbers, compress}{natbib}

\usepackage[preprint]{neurips_2020}

\usepackage[utf8]{inputenc} 
\usepackage[T1]{fontenc}    
\usepackage{url}            
\usepackage{booktabs}       
\usepackage{amsfonts}       
\usepackage{nicefrac}       
\usepackage{microtype}      
\usepackage{natbib}
\usepackage{algorithm}
\usepackage{algorithmic}
\usepackage{bbm}
\usepackage{microtype}
\usepackage{graphicx}
\usepackage{subfigure}
\usepackage{varwidth}
\usepackage{capt-of}
\usepackage{xspace}
\usepackage{enumitem}
\usepackage{caption}
\usepackage{dcolumn}
\usepackage{mathtools}
\usepackage[thinc]{esdiff}
\usepackage{graphicx}
\usepackage{mathtools}
\usepackage{pgfplots}
\usepackage{marvosym,listings,etoolbox}
\usepackage[breaklinks=true]{hyperref}
\usepackage[]{siunitx}
\usepackage{booktabs}
\robustify\bfseries
\usepackage[version=4]{mhchem}
\newcommand\diag[4]{%
  \multicolumn{1}{p{#2}|}{\hskip-\tabcolsep
  $\vcenter{\begin{tikzpicture}[baseline=0,anchor=south west,inner sep=#1]
  \path[use as bounding box] (0,0) rectangle (#2+2\tabcolsep,\baselineskip);
  \node[minimum width={#2+2\tabcolsep-\pgflinewidth},
        minimum  height=\baselineskip+\extrarowheight-\pgflinewidth] (box) {};
  \draw[line cap=round] (box.north west) -- (box.south east);
  \node[anchor=south west] at (box.south west) {#3};
  \node[anchor=north east] at (box.north east) {#4};
 \end{tikzpicture}}$\hskip-\tabcolsep}}
 
\newcommand{\sinead}[1]{{\color{red}{\bf\sf [Sinead: #1]}}}

\usepackage{amsthm,amsmath,mathtools}
\usepackage{amssymb}

\newtheorem{theorem}{Theorem}

\newtheorem{assumption}{Assumption}

\newtheorem*{remark}{Remark}

\newcommand{\alg}{{\textsc{DNND}}\xspace}

\title{A Nonparametric Bayesian Model\\ 
for Sparse Dynamic Multigraphs}

\author{
 Elahe Ghalebi\\
  TU Wien\\
  \And
Hamidreza Mahyar\\
Boston University\\
\And  
  Radu Grosu\\
  TU Wien\\
\AND  
 Graham W. Taylor\\
 University of Guelph\\Vector Institute\\
\And
 Sinead A. Williamson\\
University of Texas, Austin
}

\begin{document}
\maketitle
\begin{abstract}
  Many large networks take the form of sequences of different types of interactions between entities, which can be represented as a sparse, structured, dynamically evolving multigraph. Bayesian edge-exchangeable models have been proposed as a model for sparse multigraphs, and these have been incorporated into hierarchical models that are able to capture community-like structure. However, these models assume exchangeability of the edges, precluding us from capturing dynamic behavior, such as the tendency of individuals to reply to recent emails. 
To capture evolving graph dynamics, we propose a dynamic Bayesian nonparametric model for interaction networks that tends to reinforce recent behavioral patterns. Our Dynamic Nonparametric Network Distribution (\alg) describes a generative distribution over sequences of interactions, captured as a time-evolving mixture of dynamic behavioral patterns, that is able to capture both sparse and dense behavior. Conditioned on the hyperparameters, we are able to sample directly from the posterior distribution over the mixing components. The resulting posterior predictive distribution can be used to predict future interactions. We demonstrate impressive predictive performance against a range of state-of-the-art dynamic graph models. 
\end{abstract}
\section{Introduction}\label{sec:intro}
Many social interaction networks can be represented in terms of a multigraph---\textit{i.e.} a graph where there can be multiple edges between two vertices. For example, vertices might correspond to individuals, with each edge representing an email between two individuals. 
In large-scale applications, such multigraphs are typically sparse, with the number of edges being small relative to the number of unconnected pairs of vertices. Concretely, we call a distribution over graphs or multigraphs sparse if the number of edges grows sub-quadratically with the number of vertices. 

Edge-exchangeable models \cite{CaiCampbellBroderick2016,CraneDempsey2018} have been proposed as models for sparse multigraphs, and hierarchical variants allow the incorporation of community-type structure \cite{Williamson2016}. Such models assume that more edges and vertices will be seen in the future, making them appropriate for growing graphs, and under certain settings they are provably sparse. However, they assume that the distribution over multigraphs is stationary, and that the resulting multigraph is invariant to reordering the arrival times of the edges. In practice, most real-world networks are dynamic, with the underlying distribution evolving over time. Despite this, there is a lack of models for sparse, dynamically evolving  multigraphs.

We show that the basic edge-exchangeable framework of \cite{CaiCampbellBroderick2016} can be adapted to yield dynamically evolving multigraphs with provable sparsity, replacing the underlying distribution on the space of vertices with a time-dependent process \cite{BleiFrazier2011}. This construction allows the distribution to evolve over time, in a manner that encourages new edges to contain recently visited vertices. We incorporate this basic dynamic multigraph into a dynamic, hierarchical model that retains this sparsity while capturing complex, time-evolving interaction structure. The Dynamic Nonparametric Network Distribution (\alg) uses a temporally evolving clustering structure and a  hierarchical Bayesian nonparametric framework to capture both global changes in cluster popularity and shifting dynamics within clusters. A judicious choice of base measure for the cluster-specific distributions means that our distribution can generate either sparse or dense multigraphs, with the degree of sparsity controlled by a single parameter.  The increased flexibility allowed by our model leads to improved performance over both its exchangeable counterpart and over a range of state-of-the-art dynamic network models. 

\textbf{Contributions:} (1) We show that a sequence of edges whose distributions are governed by a distance-dependent Chinese restaurant process \cite{BleiFrazier2011} form sparse, non-stationary multigraphs---even though comparable, edge-exchangeable multigraphs based on the Chinese restaurant process are not sparse;  (2) we construct a flexible hierarchical model, \alg, appropriate for dynamically evolving, structured multigraphs. We show that this distribution can be used to model sparse or dense interaction networks, and propose an MCMC algorithm for inference; (3) we show that \alg outperforms state-of-the-art Bayesian dynamic network models over three real-world network datasets. 
\section{Background and related work}\label{sec:related}


To provide appropriate context, in this section we introduce the notion of graph sparsity (Section~\ref{subsec:graphsparsity}, and discuss existing Bayesian models for multigraphs in the stationary (Section~\ref{subsec:bg_static}) and dynamic (Section~\ref{subsec:bg_dynamic}) setting. Finally, we review the ddCRP (Section~\ref{subsec:ddCRP}), which will be used in \alg.

\subsection{Graph sparsity}\label{subsec:graphsparsity} 
The density of a graph with $E$ edges and $V$ vertices is the ratio of the number of edges to the number of potential edges. For binary, undirected graphs, this density is $2E/V(V-1)$. We can extend this notion to distributions over graphs, by looking at how the density behaves as the graph grows. We say a graph is sparse if the number of edges grows subquadratically with the number of vertices (see e.g. \cite{Orbanz:Roy:2014} for a discussion of sparsity and density in graphs).

Following \cite{CaiCampbellBroderick2016}, we extend this definition to multigraphs, saying a multigraph is sparse if the number of edges grows subquadratically with the number of vertices. Empirically, many large real-world graphs and multigraphs are sparse \cite{demaine2014structural,CaronFox2017}.

\subsection{Vertex-exchangeable and sparse-exchangeable models for multigraphs}\label{subsec:bg_static}
Most Bayesian models for (multi)graphs exhibit a form of exchangeability, meaning that they are invariant to temporal reordering of the data. As we will see in Section~\ref{subsec:bg_dynamic}, while these models are inherently unsuited for dynamically evolving graphs, most dynamic Bayesian models are based on an exchangeable counterpart. The models described in this paper are no exception.

There are several forms of exchangeability that are appropriate in the context of graphs, depending on which aspect of the graph we consider reordering; each context of exchangeability leads to different model properties.  The models described in this paper are based on edge-exchangeable graphs, which we will describe 
in Section~\ref{subsec:edge-exch}. Here, we review the two other families of exchangeable (multi)graphs.

\textbf{Vertex-exchangeable graphs} \cite{Aldous1981,Hoover:1979,Orbanz:Roy:2014} are graphs where the distribution over the adjacency matrix is invariant to jointly permuting the row and column indices. This occurs when the presence (or in the case of a multigraph, the number of instances) of an edge between vertices $u$ and $v$ is a random variable parameterized by the value of some function $\theta(u, v)$ and is conditionally independent from other edges given $\theta(u, v)$---for example, if $e_{u,v}\sim\mbox{Bernoulli}(\theta(u, v))$. This class includes the stochastic blockmodel \cite{Snijders:Nowicki:1997,Karrer:Newman:2011}; the infinite relational model \cite{irm}; the mixed-membership stochastic blockmodel \cite{airoldi2008mixed}; the latent feature relational model \cite{miller2009nonparametric}; and Poisson factor analysis \cite{ZhouCarin2013,gopalan2015scalable}. While these models are able to capture interesting community structure, the resulting graphs are dense almost surely, \cite{Aldous1981, Hoover:1979}. This makes them a poor choice for large real-world multigraphs, which are typically sparse.\looseness=-1

\textbf{Sparse exchangeable (multi)graphs} \cite{CaronFox2017,Veitch:Roy:2015,veitch2019sampling,borgs2019sampling} consider an alternative notion of exchangeability, based on increments of an underlying random measure. For example, \cite{CaronFox2017} describes a model for graphs and multigraphs based on an underlying Poisson process, whose rate measure is distributed according to a generalized gamma process \cite{brix1999generalized}; this family is generalized by \cite{Veitch:Roy:2015}. Such a construction yields sparse graphs with a power-law degree distribution, properties that are common in large social networks. \cite{lee2018bayesian} proposes an extension of such a model with community structure. Unlike the edge-exchangeable graphs that we will consider in Section~\ref{subsec:edge-exch}, sparse exchangeable graphs assume a fully observed graph where no new edges will be seen between existing vertices, making them poorly suited to prediction tasks in dynamically growing interaction networks.

\subsection{Edge-exchangeable multigraphs}\label{subsec:edge-exch}

Edge-exchangeable multigraphs \cite{CaiCampbellBroderick2016,CraneDempsey2018,Williamson2016} construct a multigraph based on an infinitely exchangeable sequence of edges. Edges are sampled from a distribution over pairs of vertices, which is typically constructed by sampling each vertex independently from some nonparametric probability distribution. Unlike vertex-exchangeable and sparse-exchangeable models, edge-exchangeable multigraphs can grow over time.
For certain choices of distribution over the vertices, edge-exchangeable multigraphs are sparse \cite{CaiCampbellBroderick2016,CraneDempsey2018}. Loosely, this occurs when the distribution is heavy-tailed, as this ensures the probability of a new edge adding a new vertex to the graph remains high enough to keep the number of edges subquadratic in the number of vertices. This holds for a class of completely random measures \cite{CaiCampbellBroderick2016} that includes the generalized gamma process \cite{brix1999generalized}, and for the Pitman-Yor process \cite{CraneDempsey2018,Pitman:Yor:1997}.

\textbf{Hierarchical edge-exchangeable multigraphs}. Edge-exchangeable multigraphs can exhibit sparsity and power-law degree distribution,  but they lack complex structure, since the two vertices that comprise each edge are chosen independently. The mixture of Dirichlet network distributions (MDND) \cite{Williamson2016} breaks this independence by using a mixture of edge-exchangeable models. Edges are associated with one of an unbounded number of clusters, with the cluster assignments distributed according to a Chinese restaurant process (CRP). Within each cluster, edges are generated according to an edge-exchangeable multigraph sequence. In this sequence, the directed edges are generated by sampling a ``sender'' and a ``recipient'' from two separate Dirichlet processes (DPs) to allow asymmetry in a directed setting which are coupled via a shared base measure. 
However, the choice of DPs as distributions over vertices mean that MDND does not yield sparse graphs \cite{CaiCampbellBroderick2016,CraneDempsey2018}. As we show in Section \ref{sec:model1}, even if we were to replace the top-level DP with a heavy tailed distribution, this would not guarantee sparsity in the resulting multigraph.

\subsection{Models for dynamic graphs}\label{subsec:bg_dynamic}

Since real-world interaction networks often evolve over time, there has been significant research attention on dynamic graph models. A common approach relies on the extensions of \emph{stationary} network models to a \emph{dynamic} framework. Although, there have been many dynamic extensions of non-Bayesian models such as the exponential random graph model~\cite{guo2007recovering} and matrix and tensor factorization-based methods~\cite{dunlavy2011temporal}, here we focus on Bayesian models relevant to the present work.


Most dynamic Bayesian networks extend vertex-exchangeable graphs. \cite{xu2014dynamic} and \cite{durante2014nonparametric} extend the stochastic blockmodel to allow time-evolving parameters. \cite{xu2015sbtm} relaxes exchangeability assumptions, allowing the presence or absence of edges to directly influence future edge probabilities. Temporal dynamics have also been added to the mixed membership stochastic blockmodel framework \cite{fu2009dynamic,xing2010state,ho2011evolving}, the infinite relational model \cite{ng2017dynamic} and the latent feature relational model \cite{foulds2011dynamic,heaukulani2013dynamic,kim2013nonparametric}. Recently, several models have extended Poisson factor analysis. The dynamic gamma process Poisson factorization (DGPPF)~\cite{acharya2015nonparametric} introduces dependency by incorporating a Markov chain of marginally gamma random variables into the latent representation. The dynamic Poisson gamma model (DPGM)~\cite{yang2018poisson} extends a bilinear form of Poisson factor analysis~\cite{zhou2015infinite} in a similar manner. The dynamic relational gamma process model (DRGPM)~\cite{yang2018dependent} also incorporates a temporally dependent thinning process. 

Much less work has been carried out on dynamic extensions of sparse-exchangeable or edge-exchangeable graphs. \cite{PallaCaronTeh2016} extends the sparse-exchangeable model of \cite{CaronFox2017} to use a time-dependent base measure, and assume edges have a geometric lifespan. In the edge exchangeable case, \cite{ng2017dynamic} incorporates temporal dynamics into the MDND by introducing a latent Gaussian Markov chain, and a Poisson vertex birth mechanism; while it offers empirical evidence of sparsity, it is not proven to yield sparse multigraphs. \cite{ghalebi2018dynamic} extends the MDND to partially observed data and use a temporally informed inference algorithm, but the underlying model is stationary. 



\subsection{Distance-dependent Chinese restaurant process}\label{subsec:ddCRP}
Edge-exchangeable graphs generate edges according to either a single exchangeable distribution over vertices, or a collection of coupled exchangeable distributions over vertices. A natural way to incorporate temporal dependence in such a model is to replace the associated exchangeable distributions with temporally varying clustering mechanisms. In this paper, we choose to use the distance-dependent CRP (ddCRP) \cite{BleiFrazier2011}, since it yields desirable sparsity results. 

Under the ddCRP with concentration parameter $\tau$ and non-negative, non-increasing decay function $f$ such that $f(\infty)=0$, the probability of an observation $x_i$ joining a cluster $k$ is,
\begin{equation}
P(z_i=k|z_{<i}) \propto \begin{cases} \sum_{j: z_j=k} f(d_{i,j}) & k\leq K_i\\
\tau & k=K_i + 1\end{cases}, ~~~ d_{i, j} = \begin{cases} t_i - t_j & t_i \geq t_j\\ \infty &
 \mbox{otherwise}\end{cases}
\label{eqn:ddcrp}
\end{equation}
where $z_i$ is the cluster assignment of the $i$-th edge, $K_i$ is the number of previously seen clusters 
 and the distance $d_{i,j}$ captures how much time has elapsed between $x_i$ and $x_j$. The concentration parameter $\tau$ controls the expected number of clusters. The decay function means that new samples are more likely to be from the same component as recently seen samples than older samples, allowing the predictive distribution to evolve over time. A known limitation of the ddCRP is that it assumes that all data has been observed up to the current time point: the distribution is not invariant to adding edges at previously observed time points. This is not a concern in our setting, since we are typically able to observe past instances of the full graph, and are interested in predicting future edges.

\section{Edge sequence distributions for sparse, dynamic multigraphs}\label{sec:basic_model}
In the context of edge-exchangeable multigraphs,
we can achieve sparsity if the underlying distribution over vertices has sufficiently heavy tails \cite{CaiCampbellBroderick2016, CraneDempsey2018}. Loosely, edge-exchangeable multigraphs obtain sparsity if the probability of incorporating a new vertex remains sufficiently large as the graph grows, so that the number of vertices is large relative to the number of edges. However, by their nature, edge-exchangeable models are not appropriate for data whose distribution varies over time, since their distributions are invariant to reordering the arrival times. In this section, we describe an alternative way of ensuring sparsity in a non-exchangeable setting, by allowing the probability of resampling an existing vertex to decay over time. 

Let us define $\mathcal{E} = (e_1, e_2,\dots)$ to be a collection of directed edges, each expressed as an ordered tuple $e_i = (s_i, r_i)$. Using the language of an email communication network, we consider an edge as originating at a sender $s_i$, and leading to a recipient $r_i$. 
In the basic edge-exchangeable multigraph \cite{CaiCampbellBroderick2016}, both sender and recipient are sampled i.i.d.\ from some discrete distribution,
$s_i, r_i \stackrel{\scriptscriptstyle{iid}}{\sim} f$.

Instead of sampling from a fixed $f$, we can sample the two end points of an edge from a dynamically evolving distribution. We choose to use the ddCRP (Equation \ref{eqn:ddcrp}), meaning that the probability of selecting vertex $v$ as either the $i$th sender, or the $i$th recipient, is
\begin{equation}
    P(s_i = v| s_{<i}, r_{<i}) = P(r_i = v| s_{<i}, r_{<i}) \propto 
    \begin{cases}\textstyle\sum_{j: s_j=v}  f(d_{i, j}) + \sum_{j: r_j=v} f(d_{i, j}) \quad &v\leq V_i \\
    \tau \quad &v = V_i +1
    \end{cases}
    \label{eqn:ddCRPgraph}
\end{equation}
where the distance $d_{i,j}$ and decay function $f$ are as described in Section~\ref{subsec:ddCRP}, and vertices are numbered in order of appearance with $V_i$ being the number of previously seen vertices.
A number of other dependent or dynamic nonparametric processes could be used in place of the ddCRP. We choose the ddCRP because, under certain mild conditions on $f$, it yields sparse multigraphs, as we show below.

\begin{assumption}
The decay function $f$ in Equation~\ref{eqn:ddCRPgraph}, and the rate of arrival of edges $e_i$, satisfies $\sum_{i<n}f(d_{i,n})\leq D n^a$ for some $D<\infty$, $a<0.5$, and all $n$.
\end{assumption}

\begin{theorem}\label{the:sparsity}
If Assumption 1 holds, then the number of edges grows subquadratically with the number of vertices, and so the multigraph is sparse.\label{thm:sparsesimple}
\end{theorem}
The proof is provided in Appendix \ref{sec:proofs}.

\section{Sparse, structured multigraphs with temporal dynamics}\label{sec:model1}
While the edge sequence distribution described in Section~\ref{sec:basic_model} is sparse, it lacks structure. In the context of edge-exchangeable graphs, MDND, described in Section~\ref{subsec:edge-exch}, brings clustering structure and incorporates asymmetry between senders and recipients, yet it sacrifices sparsity. Further, since edges in MDND are exchangeable (\textit{i.e.}~the probability of the multigraph is invariant to reordering of the edge arrival times),  MDND is not a good fit for dynamically growing multigraphs, where the underlying mechanism is non-stationary. 

Inspired both by MDND and the dynamic distribution discussed in Section~\ref{sec:basic_model}, we propose a new model, the Dynamic Nonparametric Network Distribution (\alg), for dynamic multigraphs with community structure. Like MDND, \alg clusters edges into an unbounded number of clusters, and within each cluster, has separate distributions for sender and recipient; however the nature of these distributions are modified to ensure the resulting graph is sparse and that the underlying distribution evolves over time.

The MDND uses a CRP to cluster edges. \alg replaces this with a ddCRP, meaning that the cluster probabilities can evolve over time. We also use a ddCRP (rather than a CRP) to model the distribution over edges within each cluster, so that the behavior within each cluster also evolves over time.


As with MDND, the cluster-specific distributions are coupled via a global, discrete distribution to ensure shared support. In the case of MDND, this distribution is a DP. Here, to allow sparsity in the resulting graph, we use a heavy-tailed 
distribution. We choose to use the Pitman-Yor process \cite{Pitman:Yor:1997} with concentration parameter $\gamma$ and discount parameter $\sigma$, although as we discuss later, other heavy-tailed distributions such as the generalized gamma process could also be used. The resulting distribution over directed edges $(s_i, r_i)$ takes the form
\begin{equation}
\begin{aligned}
    P(z_i = k|z_{<i}) &\propto \begin{cases} \textstyle\sum\limits_{j: z_j=k} f_1(d_{i,j}) & k \leq K_i\\ \alpha & k = K_i+1\end{cases}\qquad H  \coloneqq \sum_{i=1}^\infty h_i\delta_{\theta_i} \sim ~\text{PY}(\gamma,\sigma,\Theta)  
    \\
    s_i|z_i, s_{<i}, H &\begin{cases} = s & \mbox{w.p. }\propto \textstyle\sum\limits_{\stackrel{j: z_j=z_i}{s_j=s}}f_2(d_{i,j})\\
    \sim H & \mbox{w.p. }\propto \tau\end{cases}\qquad
    r_i|z_i, r_{<i}, H \begin{cases} = r &\mbox{w.p. }\propto \displaystyle\sum\limits_{\stackrel{j: z_j=z_i,}{r_j=r}} f_2(d_{i,j})\\
    \sim H & \mbox{w.p. }\propto \tau\end{cases}
\label{eqn:our_model}
\end{aligned}
\end{equation}

where $K_i$ is the number of previously seen clusters, $\alpha>0$, $\gamma>0$, $\tau>0$, $\sigma\in [0, 1)$, and $d_{i,j}$ is as defined in Equation~\ref{eqn:ddcrp}. $\Theta$ is some diffuse base measure on the space of vertices.
The decay functions $f_1$ and $f_2$ (see Section~\ref{subsec:ddCRP}) control the amount of influence past data points have on the current distribution over clusters and the current per-cluster distributions over vertices, respectively. For example, we might choose a window decay function $f(d) = 1[d<\lambda]$ that only allows data points within a size-$\lambda$ window to influence the current distribution, or an exponential decay function $f(d) = e^{-d/\lambda}$, where the influence decays smoothly with time. We note that, if $\sigma=0$ and $f_1(d) = f_2(d) = 1$ for all $d<\infty$,
then Equation~\ref{eqn:our_model} is equivalent to the MDND.


The parameters $\alpha$, $\gamma$, and $\tau$ have a similar effect here as in MDND: $\alpha$ governs the number of clusters; $\gamma$ governs the overall number of vertices; and $\tau$ governs the similarity between clusters. $\sigma$ controls the sparsity of the multigraph. Under the mild constraints on $f_2$  and the rate of arrival of edges described in Assumption~\ref{ass:1}, the multigraph will be sparse if $\sigma>0.5$: 
we do not require any assumptions on the global, clustering ddCRP with decay function $f_1$.

\begin{assumption}\label{ass:1}
The decay function $f_2$ in Equation~\ref{eqn:our_model} satisfies $\sum_{i=1}^j f_2(d_{i,j})\leq D$ for some $D<\infty$ and all $j$.
\end{assumption}

\begin{remark}
While this condition is stricter than the condition in Assumption 1, it is easily satisfied provided the rate of arrival of edges is bounded. For example, if $f_2$ is a window function of size $\lambda$, then $D$ is the maximum number of edges arriving in a period of length $\lambda$. If $f_2(d) = e^{-d/\lambda}$, and $m$ is the maximum number of edges arriving per unit time, then
\begin{equation*}
 \textstyle \sum_{i< j} f_2(d_{i,j}) 
\leq \sum_{i<j} e^{-\left \lfloor{d_{i,j}}\right \rfloor/\lambda }
\leq \sum_{\ell=0}^\infty m e^{-\ell/\lambda} = \frac{me^{-\lambda}}{e^{-\lambda}-1}  
\end{equation*}
where the final inequality is due to the fact that there are at most $m$ observations with  $\lfloor{d_{i, j}}\rfloor=\ell$ for $\ell=1,\dots, \infty$. If $f$ is a logistic function, $f(d) = e^{-d+\lambda} / (1+e^{-d+\lambda})$, then $\sum_{i< j} f(d_{i,j})$ is bounded above by $\sum_{\ell=0}^\infty e^{\lambda} e^{-\ell} = e^{\lambda+1}/(e-1)$.

\end{remark}

\begin{theorem}\label{the:sparsity2}
If $f_2$ satisfies Assumption 2, and if $\sigma > 0.5$, multigraphs distributed according to Equation~\ref{eqn:our_model} are sparse.\label{thm:sparseDND}
\end{theorem}
The proof is given in Appendix \ref{sec:proofs}.
We note that, while we use a Pitman-Yor process in this paper, any base measure that can be used to construct a sparse, edge-exchangeable multigraph could be used here, since the incorporation of the ddCRP components does not affect the expected number of vertices. We demonstrate the resulting sparsity empirically, along with empirical demonstration of power-law behavior, in Appendix \ref{sec:eval_sparsity}.

\textbf{Inference.} We perform inference via a MCMC algorithm that combines aspects of the ddCRP inference algorithm proposed by \cite{BleiFrazier2011} and the hierarchical Dirichlet process algorithms proposed by \cite{Teh:Jordan:Beal:Blei:2006}. Full details are provided in Appendix~\ref{sec:inference_app}.

\textbf{Complexity.} The complexity of our model is $O(E\times K)$, where $K$ is the number of clusters. The complexity of blockmodel-based models is $O(V^2 \times K)$. For sparse graphs, we have $E << V^2$. In practice, the runtimes of the different models were comparable on the datasets we considered.
\section{Experiments}\label{sec:experiments}
In this section, we address the following questions: (1) How well does \alg capture the underlying multigraph behavior to predict unseen held-out edges? and (2) How accurate is \alg in terms of forecasting future interactions, compared to state-of-the-art dynamic interaction graph models? We include code to reproduce our results within the supplementary material.

\textbf{Datasets.} We evaluated our model on three real-world temporal multigraphs; 
\\
(1) \textit{Email-Eu-core temporal network (Email-Eu)} 
\cite{snapnets} consists of all incoming and outgoing emails in a large European research institution. $986$ individuals exchange about $332$\,k separate e-mails over $803$ days. We considered the first $T=7$ months of all departments of the institute, with $120121$ 
edges and $844$ vertices, with an average monthly density ($\frac{2}{T}\sum_{t=1}^T E_t/V_t(V_t-1)$) of $0.06$. 
\\
(2) \textit{CollegeMsg network} \cite{snapnets} records private messages in an online social multigraph at the University of California, Irvine. We evaluated the performance of our model on all $T=7$ months with $1899$ vertices and $59835$ interactions over $193$ days. The average monthly density is $0.02$. 
\\
(3) \textit{Social Evolution network (SocialEv)} \cite{madan2011sensing} tracks the everyday life of 70 students within a dormitory, based on mobile phone data. We consider Bluetooth connections, calls and SMSs as interactions, yielding a multigraph with high clustering coefficient and about $1M$ events over $T=10$ months. In practice, there are $187K$ edges and $74$ vertices, with average monthly density of $0.82$. 

\textbf{Experimental settings.} 
For \alg, we considered three decay functions: (1) \textsc{Window} decay: $f(d)=1[d<\lambda]$ only considers dependency with edges that are distant at most $\lambda$ from the current edge, (2) \textsc{Exponential} decay: $f(d) = e ^{-d/\lambda}$ decays exponentially with time, and (3) \textsc{Logistic} decay: $f(d) = \frac{e ^{-d+\lambda}}{1+e ^{-d+\lambda}}$ is a smooth version of window decay. We used the same decay function for $f_1$ and $f_2$. 
We used a $\mbox{Gamma}(5, 1)$ prior for both $\alpha$ and $\gamma$, a $\mbox{Gamma}(1,1)$ prior for $\tau$, a $\mbox{Beta}(1, 1)$ prior for $\sigma$ and a $\mbox{Gamma}(50,1)$ prior for $\lambda$. Empirically, we found that varying these priors did not dramatically impact performance.

We ran all algorithms for 1000 iterations for each dataset and each time slot (month). For the baseline methods, we used the optimization or sampling methods, settings and hyperparameters described in the relevant papers. All the experiments were run on a standard desktop, an Intel Xeon with \SI{2.5}{GHz} CPU and \SI{128}{GB} RAM. We include the scripts used in these experiments with our submission, and will make these publically available after publication.

\textbf{Baselines.} 
For the held-out edge prediction task, we compared against stationary MDND, described in Section~\ref{subsec:edge-exch}. We implemented MDND using the inference algorithm in Appendix~\ref{sec:inference_app}, with $\sigma=0$ and $d_{i, j}=1$ for all $i\geq j$. We found that this algorithm gave comparable results to the implementation of \cite{Williamson2016}.
For the graph forecasting task, we also compared against three recent Bayesian dynamic network models, introduced in Section~\ref{subsec:bg_dynamic}: the dynamic relational gamma process model (DRGPM) \cite{yang2018dependent}, the dynamic Poisson gamma model (DPGM) \cite{yang2018poisson}, and the dynamic gamma process Poisson factorization (DGPPF) \cite{acharya2015nonparametric}. These models are not applicable to the held-out edge prediction task, since they assume all vertices are observed. However, they can be used to predict the entire graph at the next time slot over seen vertices. We modify this distribution to be appropriate to our forecasting task by predicting the $n_T$ edges with the highest probability at time slot $T$.

\textbf{Evaluation metrics.}
We consider four evaluation metrics: Log likelihood of held-out data, F1 score for predicted future interactions, hits ratio at $k$ (hits@$k$) and average precision at $k$ (AP@$k$) for predicted future interactions. 
The held-out log likelihood allows us to evaluate whether our model is a good fit for the data. We use this to evaluate whether incorporating time-dependence and sparsity allows us to better capture variation in the data compared to MDND. Estimating test set log likelihood can be tricky in models where we have latent variables for each data point, since the likelihood depends heavily on the assignments of those latent variables, and the state space of assignments is too large to explore exhaustively. We estimate the log predictive likelihood using the ``left to right'' algorithm \cite{Wallach:2009:EMT:1553374.1553515} explained in Appendix~\ref{app:results}.

While log likelihood allows us to compare models, it does not provide a quantitative measure of how much better a model will perform on a concrete prediction task for future observations. To assess this, we consider two metrics for evaluating future predictions when we have all observations before time $T$, and want to predict the edges arriving at time $T$. To allow comparison with vertex-exchangeable models, which predict the entire adjacency matrix for a single time step, we set the time stamp for all test set edges to the time of the first test set edge. We assume that we know the total number, $N_{\mbox{\small{test}}}$, of edges in the test set, allowing us to return a predicted set of appropriate size, along with their probability of appearance. For comparison methods, we selected the $N_{\mbox{\small{test}}}$ edges with highest probability of appearing.

The F1 score---calculated as the harmonic mean of precision and recall---gives a general measure of the accuracy of the predictor, assigning equal importance to Type I and Type II errors. Average precision at $k$ looks at the proportion of the held-out edges that appear in the top $k$ predictions, averaged over samples and hits ratio at $k$ is the proportion of correct predicted edges in top-$k$ ranked edges \cite{yang2012top}. 
This allows us to dig deeper into the model's performance, considering whether the edges assigned high probability under the model are correct. The higher the F1 on the test set, the better.

\begin{table*}[t]
\centering
\caption{\small{Predictive log likelihood of held-out edges on three real-world datasets (mean $\pm$ standard deviation over time slots).}}\label{tbl:log_lhood}
\vspace{-.4cm}
\medskip
\footnotesize
\begin{tabular}{
  c|
  S[table-format=5.1(4), detect-weight, detect-shape, detect-mode]
  S[table-format=5.1(4), detect-weight, detect-shape, detect-mode]
  S[table-format=5.1(4), detect-weight, detect-shape, detect-mode]
  S[table-format=5.1(4), detect-weight, detect-shape, detect-mode]
 }
\toprule
\diag{.068em}{1.5cm}{\scriptsize{Dataset}}{\scriptsize{Method}}
&{\textsc{MDND}} &{\textsc{\alg-Window}} &{\textsc{\alg-Logistic}} &{\textsc{\alg-Exponential}} \\
\midrule
\text{Email-Eu}  &  -26868.9  \pm 375.0 & -25270.8 \pm 422.9 &  -2288.4 \pm 469.0 &  \bfseries -22260.2 \pm 399.9\\
\text{SocialEv}  &  -1181.8 \pm 152.4 &   -709.4 \pm 84.1  & -749.9 \pm 90.6  & \bfseries -699.7 \pm 84.4\\
\text{CollegeMsg}&  -4416.6 \pm 227.8 & \bfseries -3757.9 \pm 213.3  & -5041.9 \pm 375.7  & -3882.0 \pm 233.0\\
\bottomrule
\end{tabular}
 \vspace{-.4cm}
\end{table*}
\begin{figure*}[t!]
	\centering
	\subfigure[Email-Eu\label{subfig:email}]{\includegraphics[width=.3\textwidth,trim=10mm 15mm 10mm 10mm]{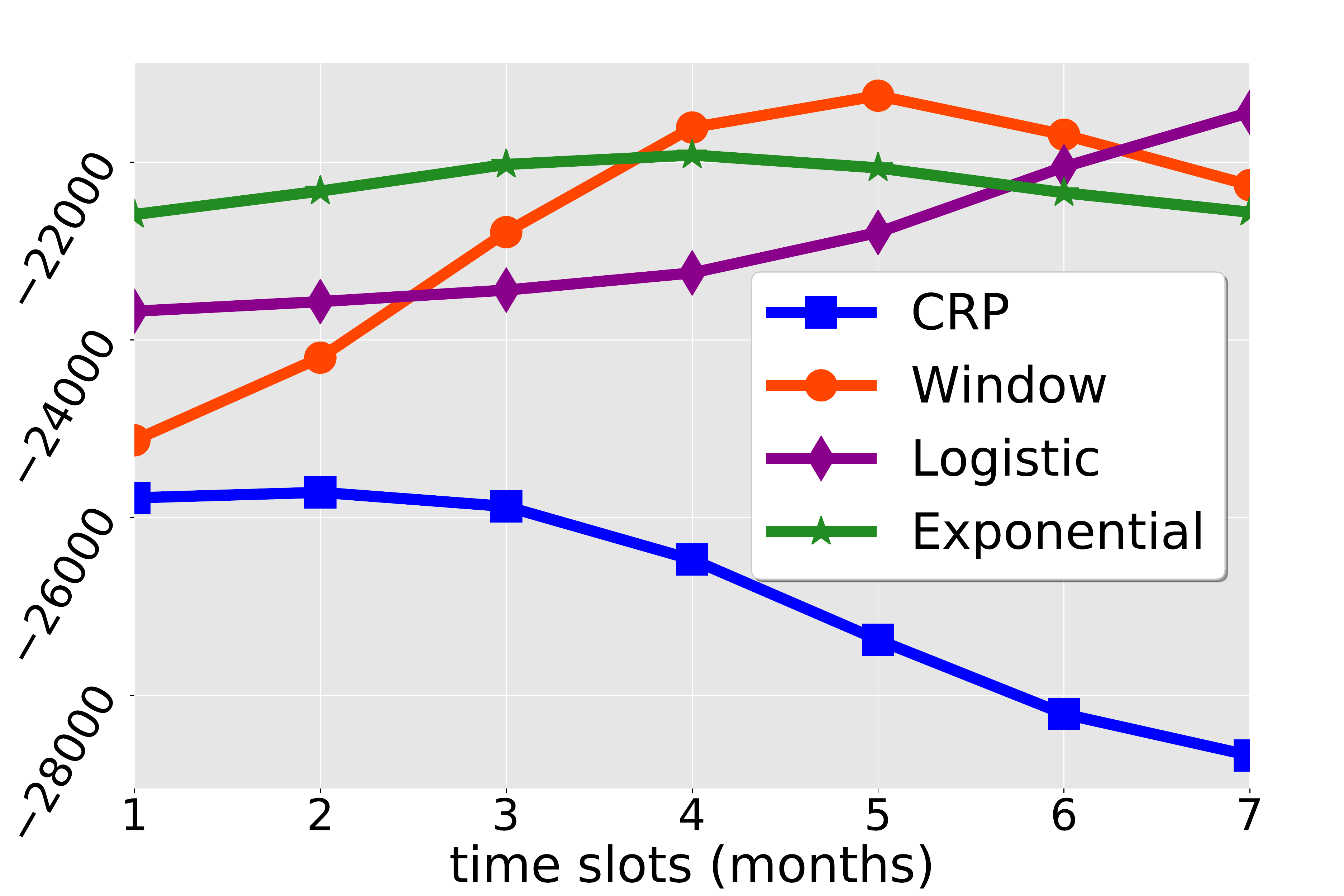}}
    \subfigure[SocialEv\label{subfig:social}]{\includegraphics[width=.3\textwidth,trim=10mm 15mm 10mm 10mm]{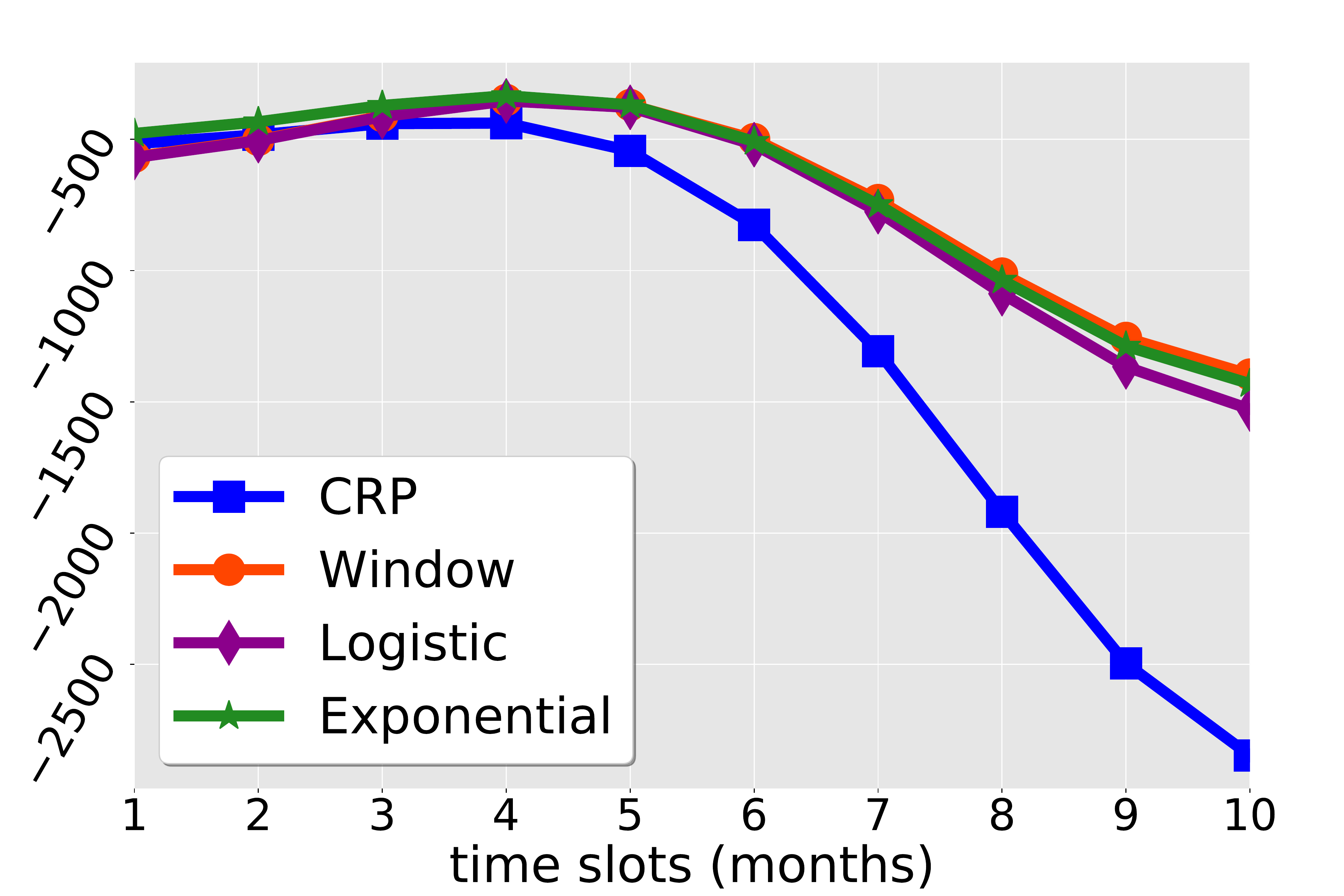}}
    \subfigure[CollegeMsg\label{subfig:college}]{\includegraphics[width=.3\textwidth,trim= 10mm 15mm 10mm 10mm] {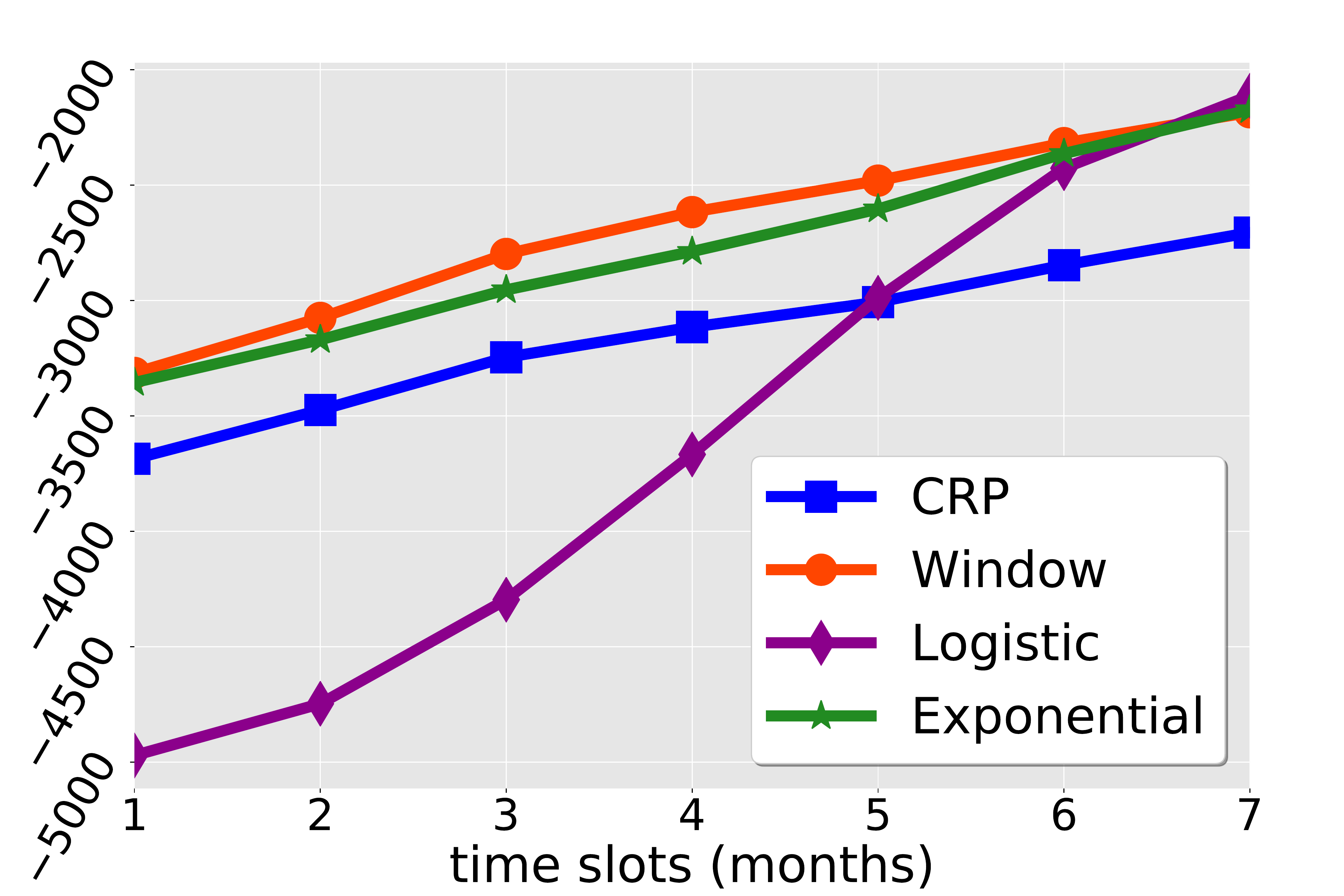}}
\vspace{-.2cm}
\caption{\small{Predictive log likelihood vs.~time slots. Each evaluation is the average value over 10 runs.}}
\label{fig:AvgPredLike}
\vspace{-3mm}
\end{figure*}

\begin{table*}[t]
\hfill
\parbox{\textwidth}{
\caption{\small{hits@$\mathtt{k}$ for the future interaction prediction task (mean value over samples).}\vspace{-.2cm}}
\centering
\scriptsize
\resizebox{1\textwidth}{!}{
\begin{tabular}{|c|ccc|ccc|ccc|ccc|ccc|}
	\toprule
	& \multicolumn{3}{|c|}{\alg} &\multicolumn{3}{|c|}{MDND} & \multicolumn{3}{|c|}{DRGPM} & \multicolumn{3}{|c|}{DPGM}& \multicolumn{3}{|c|}{DGPPF}\\ 
	\midrule
	\textit{hits@$\mathtt{k}$} & @10 & @20 & @50& @10 & @20 & @50 & @10 & @20 & @50 & @10 & @20 & @50 & @10 & @20 & @50 \\ 
	\midrule
	\textsc{Email-Eu} & \textbf{0.96} & 0.92 & \textbf{0.99}  & 0.94 & \textbf{0.96} & 0.96 & 0.3 & 0.21 & 0.23 & 0.28 & 0.21 & 0.25 & 0.32 & 0.2 & 0.2\\
	\textsc{SocialEv} & \textbf{0.78} & \textbf{0.98} & \textbf{1.0}  & 0.74 & 0.91 & \textbf{1.0} & 0.21 & 0.21 & 0.21 & 0.16 & 0.16 & 0.1 & 0.36 & 0.36 & 0.35\\
	\textsc{CollegeMsg} & \textbf{0.62} & \textbf{0.78} & \textbf{0.98}  & 0.5 & 0.63  & 0.96 & 0.06  & 0.06  & 0.06 & 0.14 & 0.15 & 0.14  & 0.17 & 0.17  & 0.16 \\
	\hline
\end{tabular}}
\label{tbl:hits}\vspace{.1cm}
}
\hfill
\parbox{\textwidth}{
\caption{\small{AP@$\mathtt{k}$ for the future interaction prediction task (mean value over samples).}\vspace{-.2cm}}
\centering
\scriptsize
\resizebox{1\textwidth}{!}{
\begin{tabular}{|c|ccc|ccc|ccc|ccc|ccc|}
	\toprule
	& \multicolumn{3}{|c|}{\alg} &\multicolumn{3}{|c|}{MDND} & \multicolumn{3}{|c|}{DRGPM} & \multicolumn{3}{|c|}{DPGM}& \multicolumn{3}{|c|}{DGPPF}\\ 
	\midrule
	\textit{AP@$\mathtt{k}$} & @10 & @20 & @50& @10 & @20 & @50 & @10 & @20 & @50 & @10 & @20 & @50 & @10 & @20 & @50 \\ 
	\midrule
	\textsc{Email-Eu} & \textbf{0.73} & \textbf{0.43} & \textbf{0.18}  & 0.66 & 0.41 & 0.18 & 0.21 & 0.14 & 0.10 & 0.16 & 0.14 & 0.12 & 0.22 & 0.17 & 0.10\\
	\textsc{SocialEv} & \textbf{0.25} & \textbf{0.16} & \textbf{0.13}  & 0.17 & 0.10 & 0.04 & 0.06 & 0.03 & 0.01 & 0.05 & 0.02 & 0.01 & 0.06 & 0.03 & 0.01\\
	\textsc{CollegeMsg} & \textbf{0.64} & \textbf{0.39} & \textbf{0.21}  & 0.43 & 0.33  & 0.2 & 0.09  & 0.05  & 0.02 & 0.1 & 0.05 & 0.02  & 0.1 & 0.05  & 0.02 \\
	\hline
\end{tabular}
}\label{tbl:aps}
}\vspace{-.4cm}
\end{table*}
\begin{figure*}[t]
	\centering  
	\subfigure[Email-Eu\label{subfig2:email}]{\includegraphics[width=.3\textwidth]{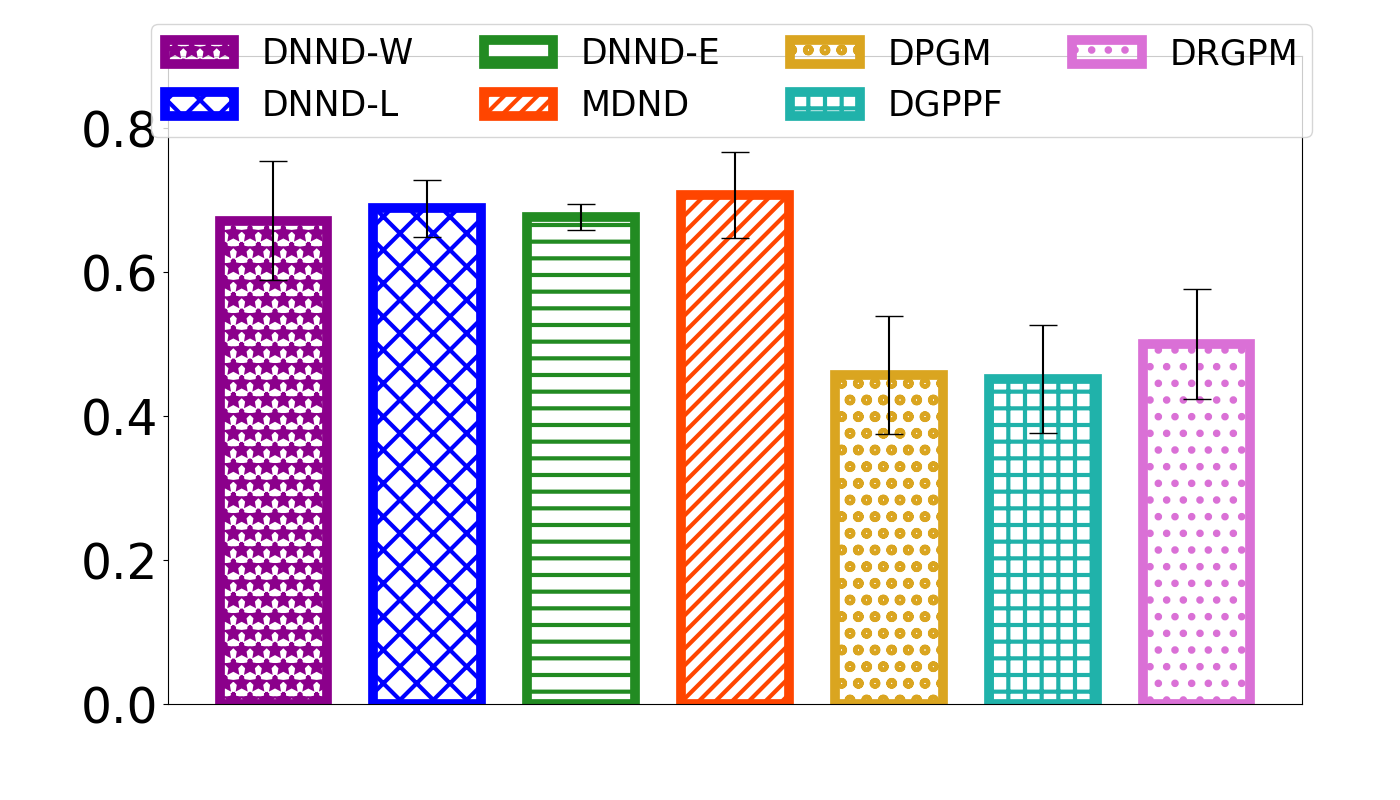}}
	\subfigure[SocialEv\label{subfig3:social}]{\includegraphics[width=.3\textwidth]{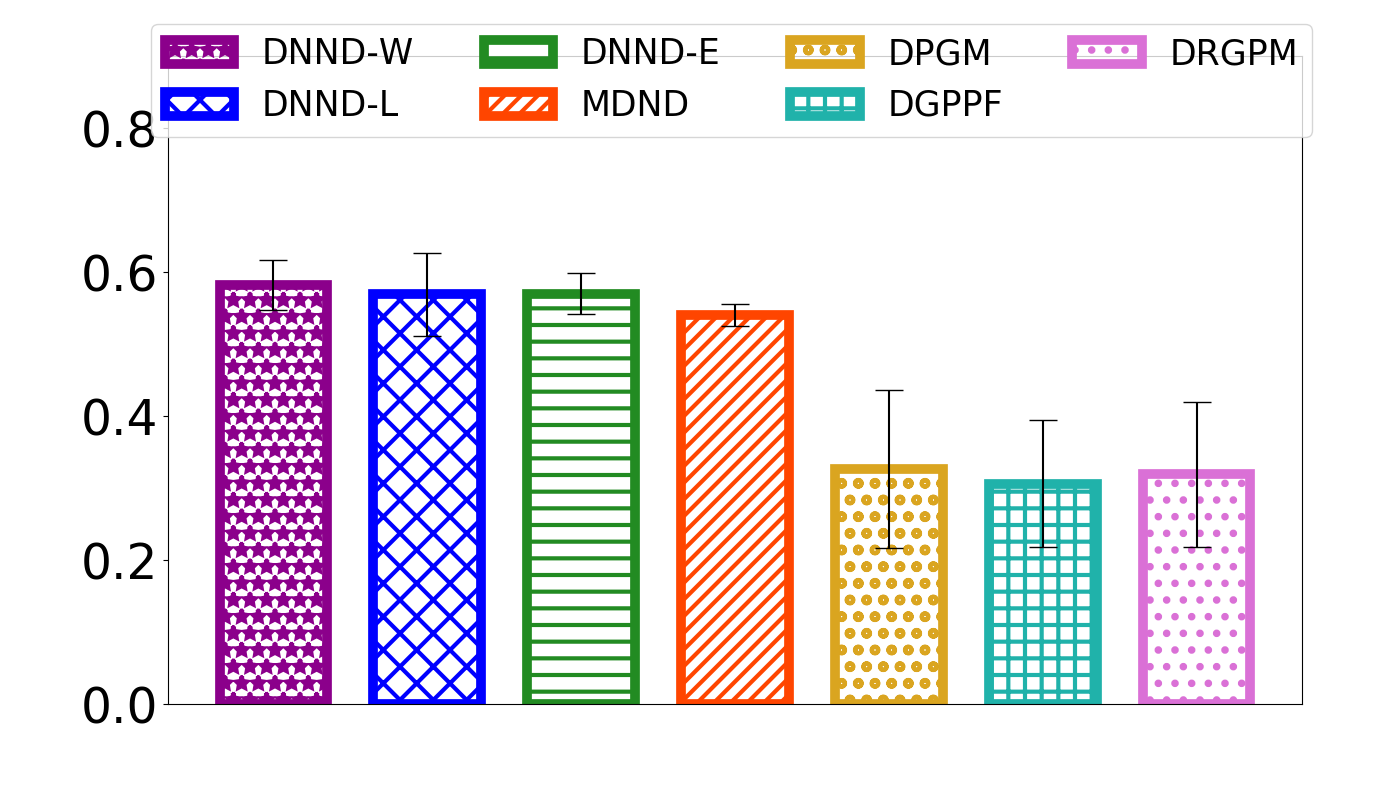}}
	\subfigure[CollegeMsg\label{subfig1:college}]{\includegraphics[width=.3\textwidth] {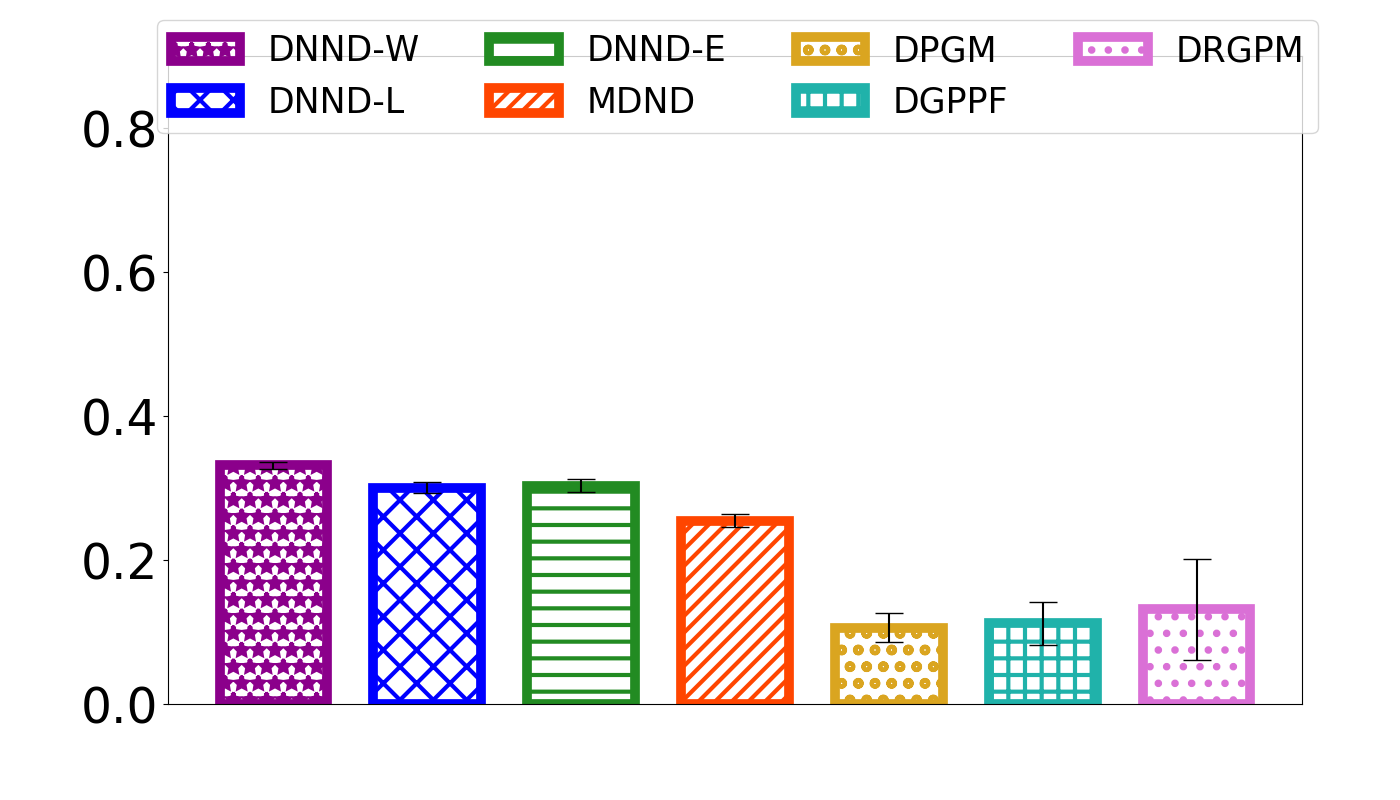}}
	\vspace{-.2cm}
	\caption{\small{F1 score for future interaction prediction. Decay functions for \alg are \textsc{Window} (W), \textsc{Exponential} (E), and \textsc{Logistic} (L) (averaged over time slots).}}
\label{fig:log-f1}
\vspace{-3mm}
\end{figure*}

\subsection{Prediction of held-out edges} 
We consider how the test-set log likelihood varies between \alg and MDND, which can be seen as a special case of \alg where $\sigma=0$ and the ddCRPs reduce to CRPs, and explore how the choice of decay function affects performance. We split each data set into time slots (one month for all datasets), and train on $85\%$ of the interactions in each time slot.

Table~(\ref{tbl:log_lhood}) shows the predictive log likelihood computed by \alg using three different decays (\textsc{Window}, \textsc{Exponential} and \textsc{Logistic}) in comparison to the CRP decay function used in \cite{Williamson2016} on three real multigraphs. Values for each dataset and each decay function represent the mean value of log likelihoods across time slots. We see that in each case, all three dynamic \alg multigraphs outperform the stationary MDND. 
In Figure~(\ref{fig:AvgPredLike}), we illustrate the predictive log likelihood per time slot, separately. Again, we see that \alg consistently outperforms the stationary MDND model across all time slots in terms of the predicted log likelihood of held-out edges. The best-performing decay function differs across datasets, likely due to different forms of temporal variation. This demonstrates that considering important properties observed in real-world data (\textit{i.e.} time dependency and sparsity) results in a better model fit. 

\subsection{Forecasting future interactions}
We evaluated the F1 score, average precision at $k$ and hits ratio at $k$ using networks of time slot $T$ as the training set and testing by predicting the network edges at time slot $T+1$. To allow comparison with vertex-exchangeable methods, we assumed in all cases that all test-set observations arrive at the same time as the first test-set observation. To predict each interaction, we estimate its likelihood based on 10 posterior samples and we report the mean value over these samples for each interaction.

Tables~(\ref{tbl:hits}) and (\ref{tbl:aps})  show the average precision and hits ratio at $k$ for \alg and the four comparative methods. Due to limited space, the numbers reported for \alg represent the best result out of the three decay functions; complete tables are included in Appendix~\ref{app:results}. Figure~(\ref{fig:log-f1}) summarizes the corresponding F1 scores. The results show that \alg performs comparably or better on both metrics across all datasets. We hypothesize that this is due to several reasons. First, \alg is explicitly designed in terms of a predictive distribution over edges, making it well-suited to predicting future edges. Second, \alg is able to increase the number of vertices over time, and is likely better able to capture natural multigraph growth. Conversely, the other methods assume the number of vertices is fixed and known---and explicitly incorporate the absence of edges at earlier time points into the likelihood. Third, unlike the other methods, \alg allows us to capture sparsity. If the underlying multigraph is sparse, then this should lead to a more accurate model.

\section{Discussion}
We have demonstrated that edge-based multigraphs constructed using the ddCRP are sparse for certain parameter settings, and have proposed a new distribution for sparse, temporally varying, structured networks based on these edge-based multigraphs. As we saw in Section~\ref{sec:experiments}, these properties translate into impressive predictive performance compared with state-of-the-art Bayesian models. 
While we focus here on the ddCRP, we note that  alternative dependent nonparametric priors could also be used. For example, the P\"{o}lya urn-based model of \cite{caron2007generalized} randomly deletes either previously seen clusters or observations, and could yield sparse graphs under appropriate deletion schemes. 
An interesting avenue for future research would be to explore alternative forms of dependency, and incorporate mechanisms that can capture link reciprocity \cite{blundell2012modelling}.

An alternative path towards sparse, dynamic graphs would be to use a dependent nonparametric process with heavy-tailed marginals, such as a dependent Pitman-Yor process \cite{sudderth2009shared}. While replacing the distribution in a simple edge-exchangeable multigraph with such a distribution would yield sparse graphs, this sparsity would not necessarily carry over to hierarchical models such as \alg. 

\section*{Broader impacts}
Our work joins a growing body of research that aims to explain the complex connectivity structure of real graph data. By incorporating sparsity and temporal dynamics in a nonparametric Bayesian framework, we provide tools that can improve link prediction and structure elucidation in interaction networks. We hope that this model will inspire researchers to use Bayesian modelling in their work and help facilitate the modeling of more complex and diverse graph-structured data.

While we have shown good predictive performance on real-world data, it is important to remember that this is not the only metric on which an algorithm should be judged. While interaction networks, such as email networks and social networks, have clear communication benefits, they also pose potential risks to users. Like most machine learning algorithms, graph algorithms risk leaking private information about users. While a Bayesian approach provides some protection against such leakage \cite{DimNelZhaMitRub2017}, our algorithm is not designed to protect user privacy. We refer the reader to \cite{BorChaSmi2015} for a discussion of privacy in sparse graphs.

Two other concerns appropriate to interaction networks---which often involve sensitive user information, and can impact the lives and decisions of their users---are the interpretability of the algorithm, and the fairness of its predictions. Since our model is based on a hierarchical clustering model, its latent structure is naturally fairly interpretable, allowing the user to explain predictions in terms of these latent clusters.

While we do not explore the fairness of our predictions in this work, it is known that the ``filter bubble'' effect can cause systemic bias \cite{MasWilYanTanEsf2020}. This occurs when a link prediction algorithm predicts that a user will interact with similar users, and so only suggests such connections. As a result, groups tend to become segregated. If, for example, men are significantly more likely to belong to a bubble including influential individuals, this puts them at a clear advantage. The clustering behavior underlying our model means that, if it were used to recommend connections, it could lead to such a filter bubble. We do not, therefore, recommend that our algorithm be the sole method used for recommending interactions to individuals, and care should be taken to avoid such undesirable consequences.


\bibliography{references.bib}
\bibliographystyle{plainnat}

$ $
\newpage
\onecolumn{}
\appendix
\textbf{\Large Supplementary material}

\section{Proofs}\label{sec:proofs}
\textbf{Theorem \ref{the:sparsity}, restated}. If Assumption 1 holds, then the number of edges grows subquadratically with the number of vertices, and so the multigraph is sparse.

\begin{proof}
Consider first samples $x_1, x_2, \dots$ from a ddCRP with decay function $f$ and concentration parameter $\tau$, such that $\sum_{i<n}f(d_{i,n})\leq D n^a$ for some $D<\infty$ and all $n$. The probability that $x_i \neq x_j$ for all $j<i$ is $\frac{\tau}{\tau + \sum_{j<i}f(d_{i,j})}$, therefore, the expected number of distinct values, $K_n$, in $n$ observations is
$$
    \mathbb{E}[K_n] = \sum_{i=1}^n \frac{\tau}{\tau + \sum_{j<i}f(d_{i,j})}\geq \frac{n\tau}{\tau + Dn^{a}}.\ 
$$
If we use this ddCRP to construct a multigraph as described in Section~\ref{subsec:ddCRP}, then the expected number of distinct vertices $V_n$ in $E_n$ edges will be
$$
\begin{aligned}
    \mathbb{E}[V_n] =&\sum_{i=1}^{E_n} \frac{\tau}{\tau + 2 \sum_{j<i}f(d_{i,j})} 
    + \sum_{i=1}^{E_n}\frac{\tau}{\tau + 2\sum_{j<i}f(d_{i,j}) + f(d_{i,i})}\\
    \geq& \frac{2E_n\tau}{\tau + 2DE_n^{a}} \sim O(E_n^{1-a}) .\
    \end{aligned}
$$
Therefore, the multigraph will be sparse provided $a<0.5$.
\end{proof}

\textbf{Theorem \ref{the:sparsity2}, restated}.
If $f_2$ satisfies Assumption 2, and if $\sigma > 0.5$, multigraphs distributed according to Equation~\ref{eqn:our_model} are sparse.

\begin{proof}
Within each cluster, senders and recipients are assigned to components according to two cluster-specific ddCRPs. If these ddCRPs were independent of one another, then each component would be associated with a unique vertex (as in the model described in Section~\ref{sec:basic_model}). Following Theorem~1 with $a=1$, by linearity of expectation, the expected total number of components would be linear in the number of edges. 

However, the ddCRPs are not independent; instead they are coupled via the discrete base measure $H$. This means that, rather than sampling a vertex from some diffuse distribution, each component samples a vertex from $H$---meaning that multiple components are associated with the same vertex. As a result, the distribution over the number of distinct vertices per edge is equivalent, up to a multiplicative constant, to that of the edge-exchangeable graph of \citet{CraneDempsey2018}, who prove that the resulting graph is sparse. We restate their argument below:

Since $H$ is distributed according to a Pitman-Yor process, the number of distinct values associated with $k$ samples from $H$ (and hence, the number of edges with degree $k$ in a Pitman-Yor edge exchangeable model) is \cite{Pitman2002}
\begin{equation}
    \frac{\Gamma(\gamma + k + \sigma)\Gamma(\gamma+1)}{\sigma\Gamma(\gamma+k)\Gamma(\gamma+\sigma)} - \frac{\gamma}{\sigma} \simeq \frac{\Gamma(\gamma+1)}{\sigma\Gamma(\gamma+\sigma)}k^\sigma \, .\label{eqn:py_num_clusters}
\end{equation}

Since, in our model,  the number of components grows linearly with the number of edges $E_n$, by the law of iterated expectations the expected number distinct vertices $V_n$ grows as $O(E_n^\sigma)$ for $\sigma>0$. Therefore, the multigraph will be sparse provided $\sigma>0.5$.
\end{proof}

\section{Empirical evaluation of sparsity and power-law degree distribution}\label{sec:eval_sparsity}
\begin{figure*}[ht]
\centering
  \includegraphics[width=.4\textwidth] {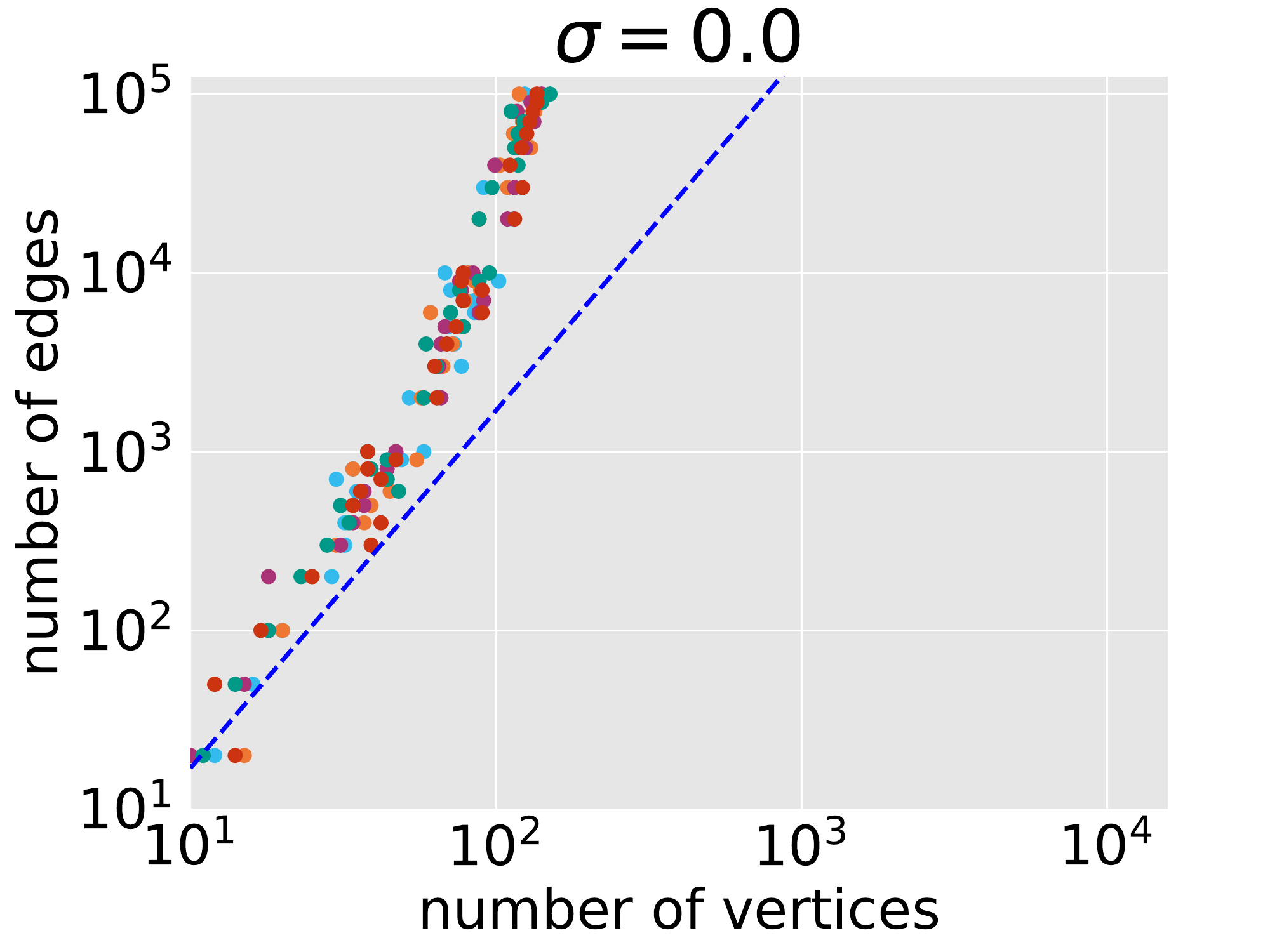} 
  \includegraphics[width=.4\textwidth] {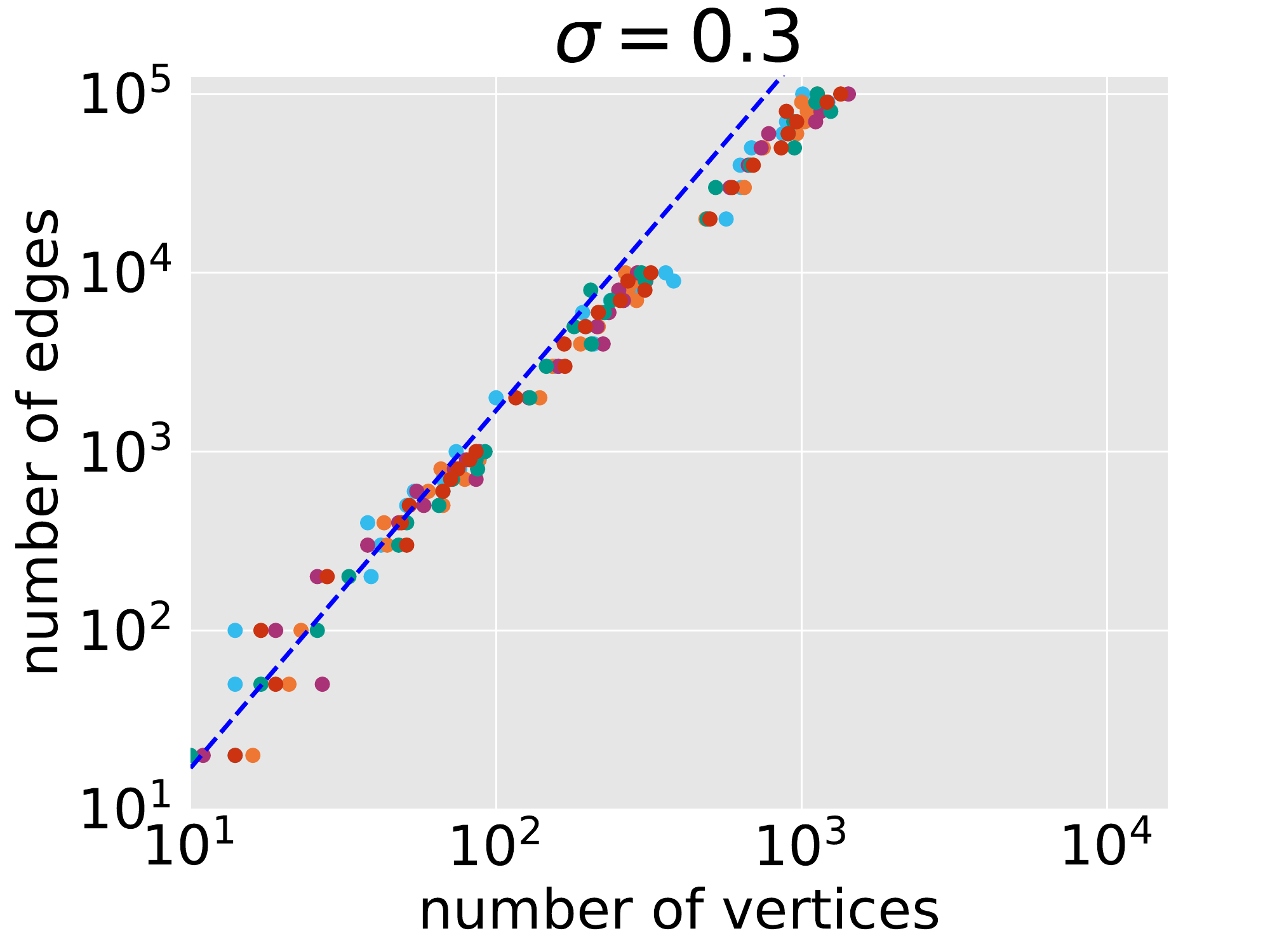} \\
  \includegraphics[width=.4\textwidth] {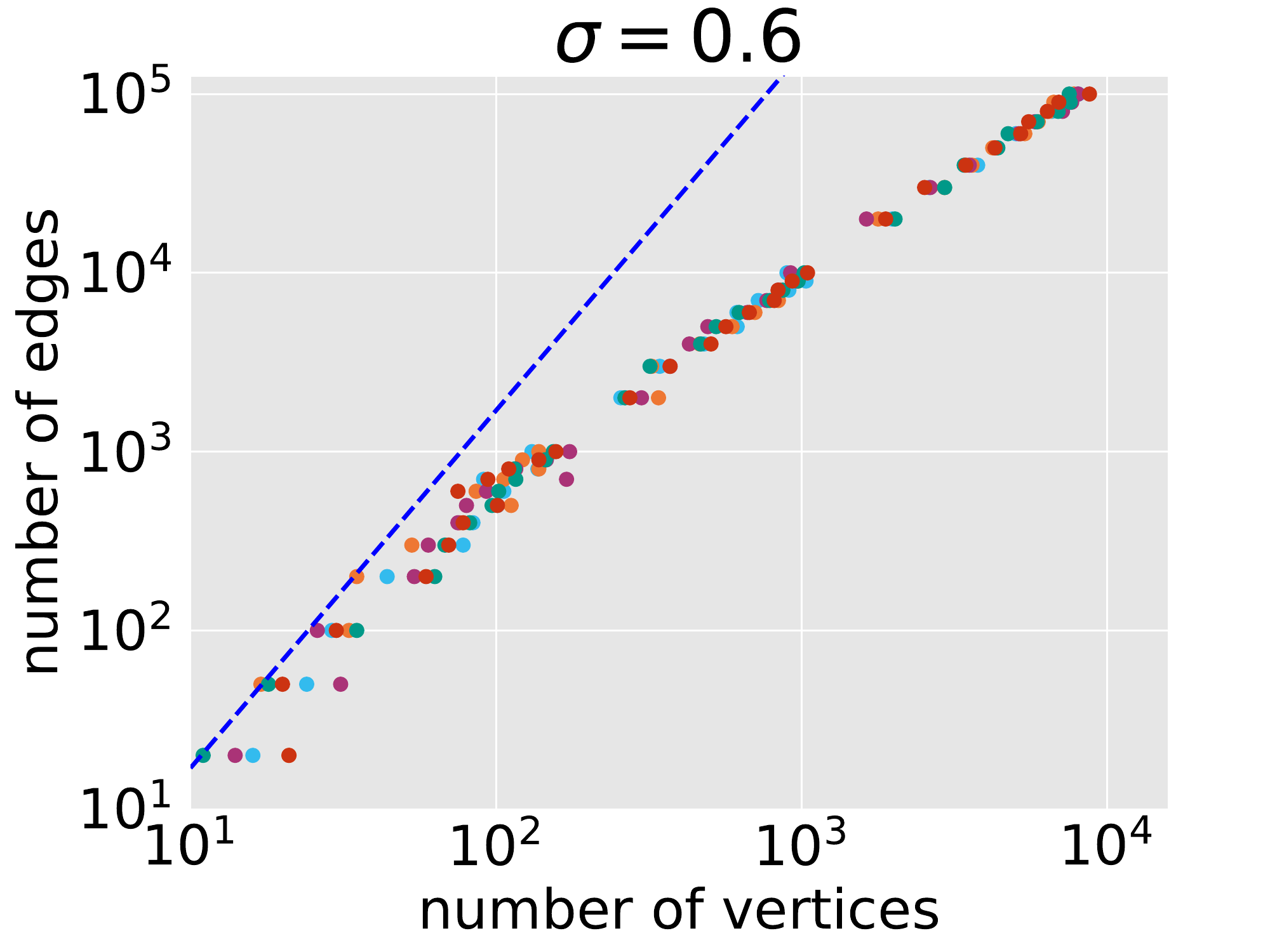} 
  \includegraphics[width=.4\textwidth] {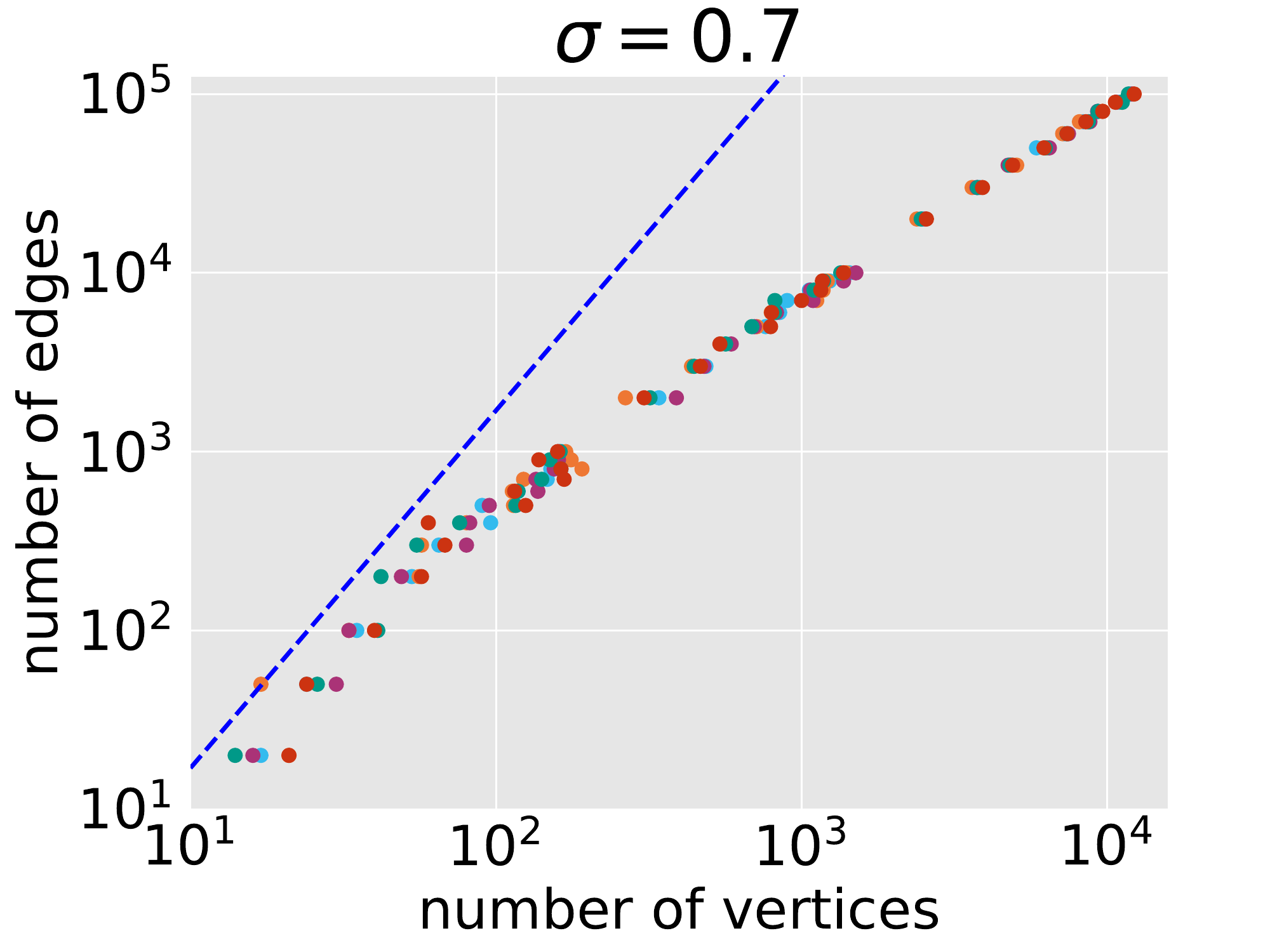}
\caption{\small{Relationship between the number of edges and the number of vertices in \alg multigraphs generated according to Equation~(\ref{eqn:our_model}), for various values of $\sigma$. Plots are shown on a log-log scale. Different colors correspond to different random seeds. The blue dashed line has a slope of 2, indicating a quadratic relationship. We see that the multigraphs become increasingly sparse as $\sigma$ increases.}}\label{fig:multigraph}
\vspace{-3mm}
\end{figure*}

In Figure~(\ref{fig:multigraph}), we empirically investigate the effect of $\sigma$ on the sparsity of multigraphs generated according to Equation~(\ref{eqn:our_model}). Recall that a model is considered sparse if the number of edges grows subquadratically with the number of vertices. Figure~(\ref{fig:multigraph}) plots the number of edges and number of vertices (on a log-log scale) for multigraphs with $\alpha=1$, $\gamma=1$, $\tau=0.2$, and various values of $\sigma$. Each dot represents a single sampled multigraph, and the blue dashed line has slope 2---providing the boundary between sparse and dense graphs. We see that, for $\sigma>0.5$, the number of edges grows subquadratically with the number of vertices, as expected according to Theorem~\ref{thm:sparseDND}. As we decrease $\sigma$, the graphs become denser. With $\sigma=0.3$, the number of edges is approximately quadratic in the number of vertices. With $\sigma=0$, corresponding to a DP-distributed base measure, the number of edges is superquadratic in the number of vertices.

\begin{figure*}[ht]
\centering
  \includegraphics[width=.4\textwidth] {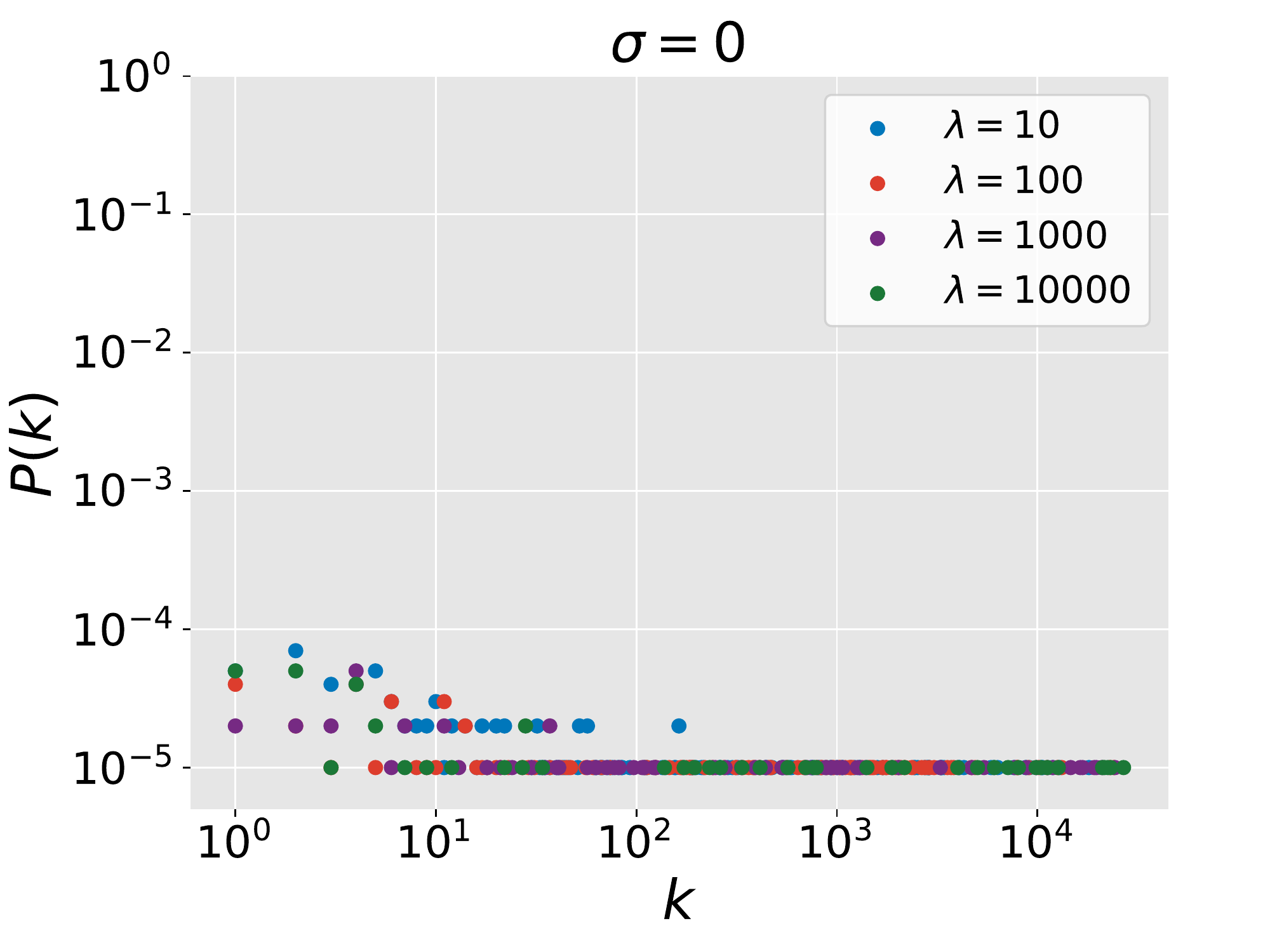}
  \includegraphics[width=.4\textwidth] {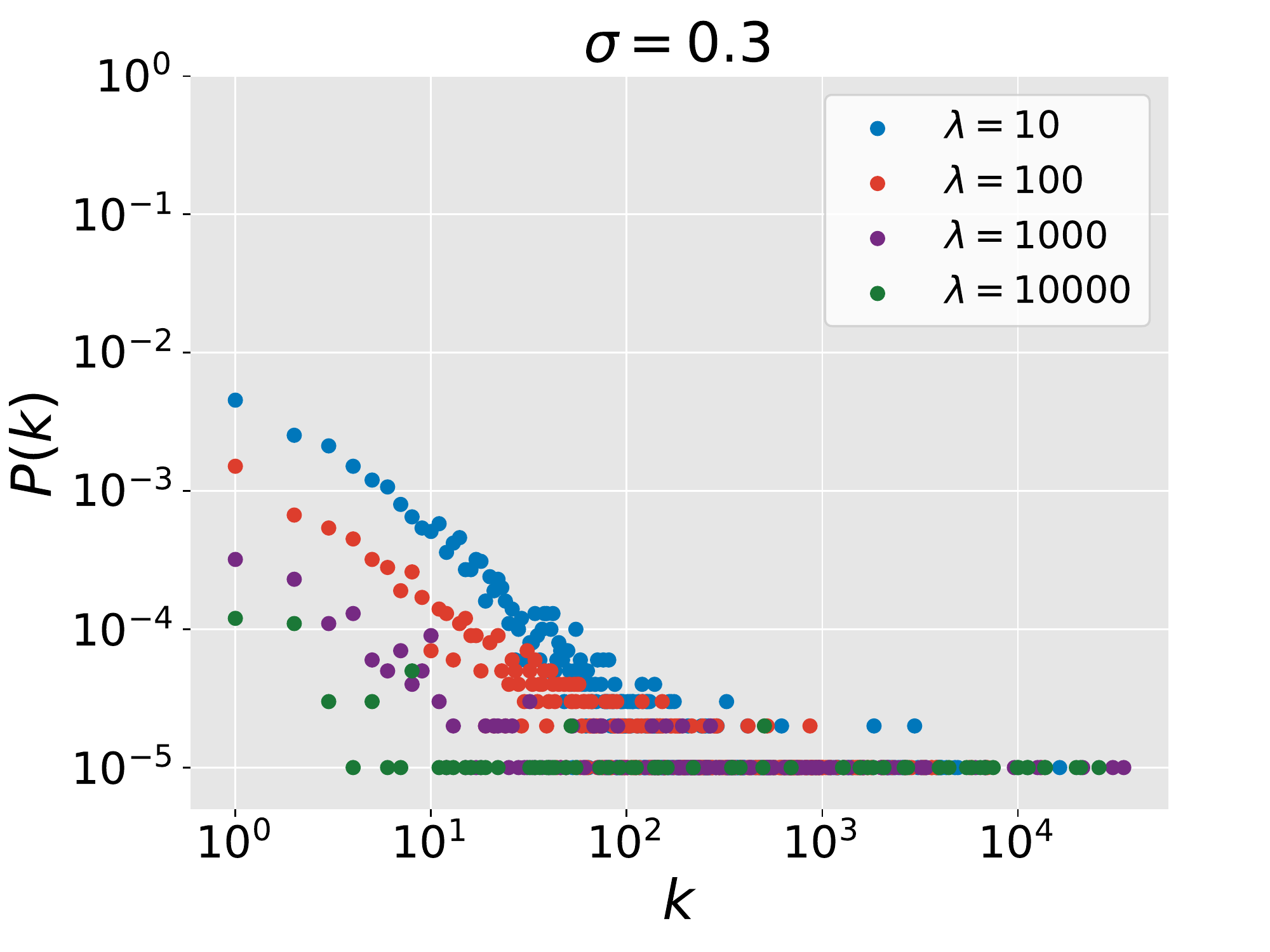} \\
  \includegraphics[width=.4\textwidth] {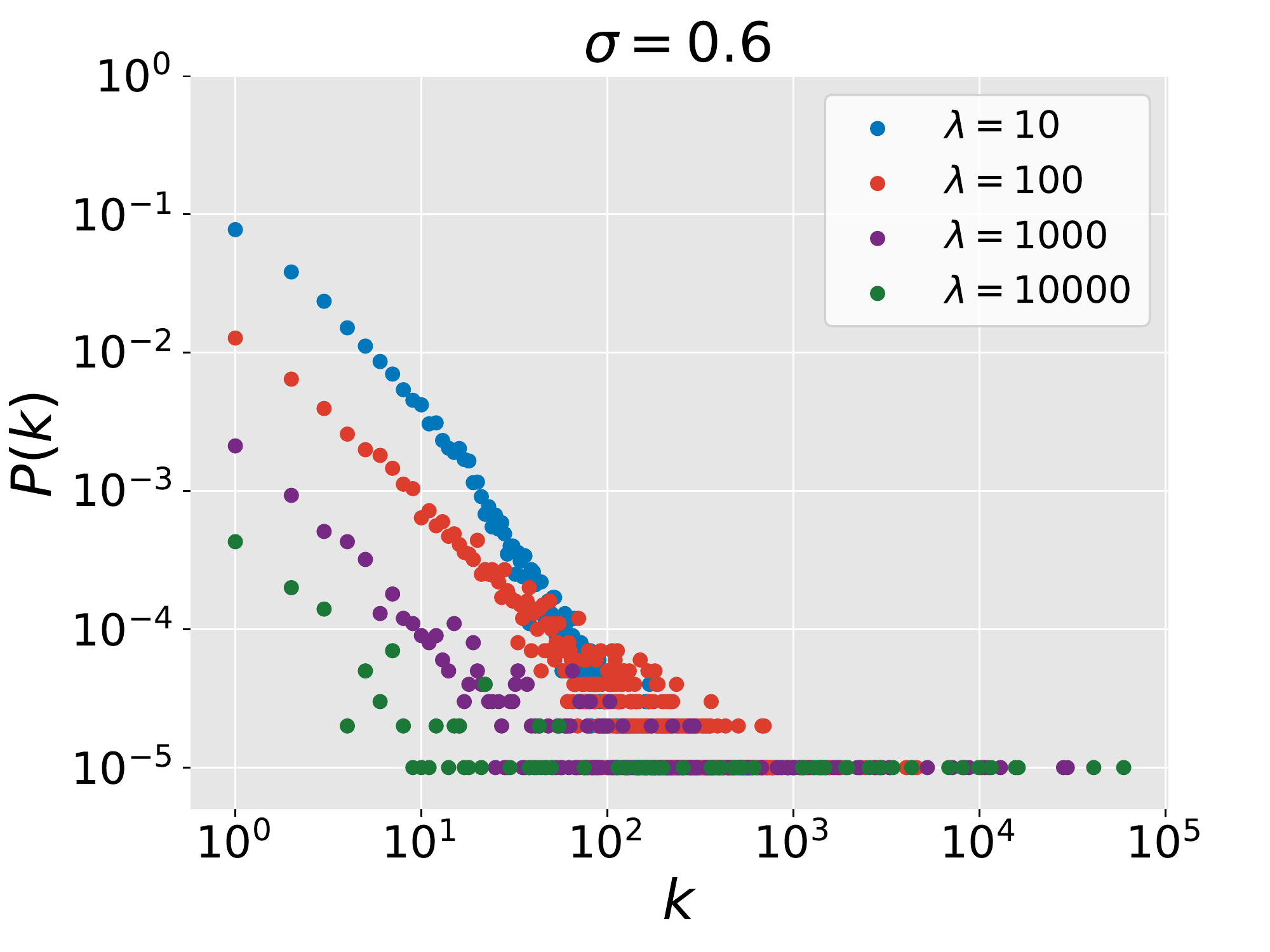} 
  \includegraphics[width=.4\textwidth] {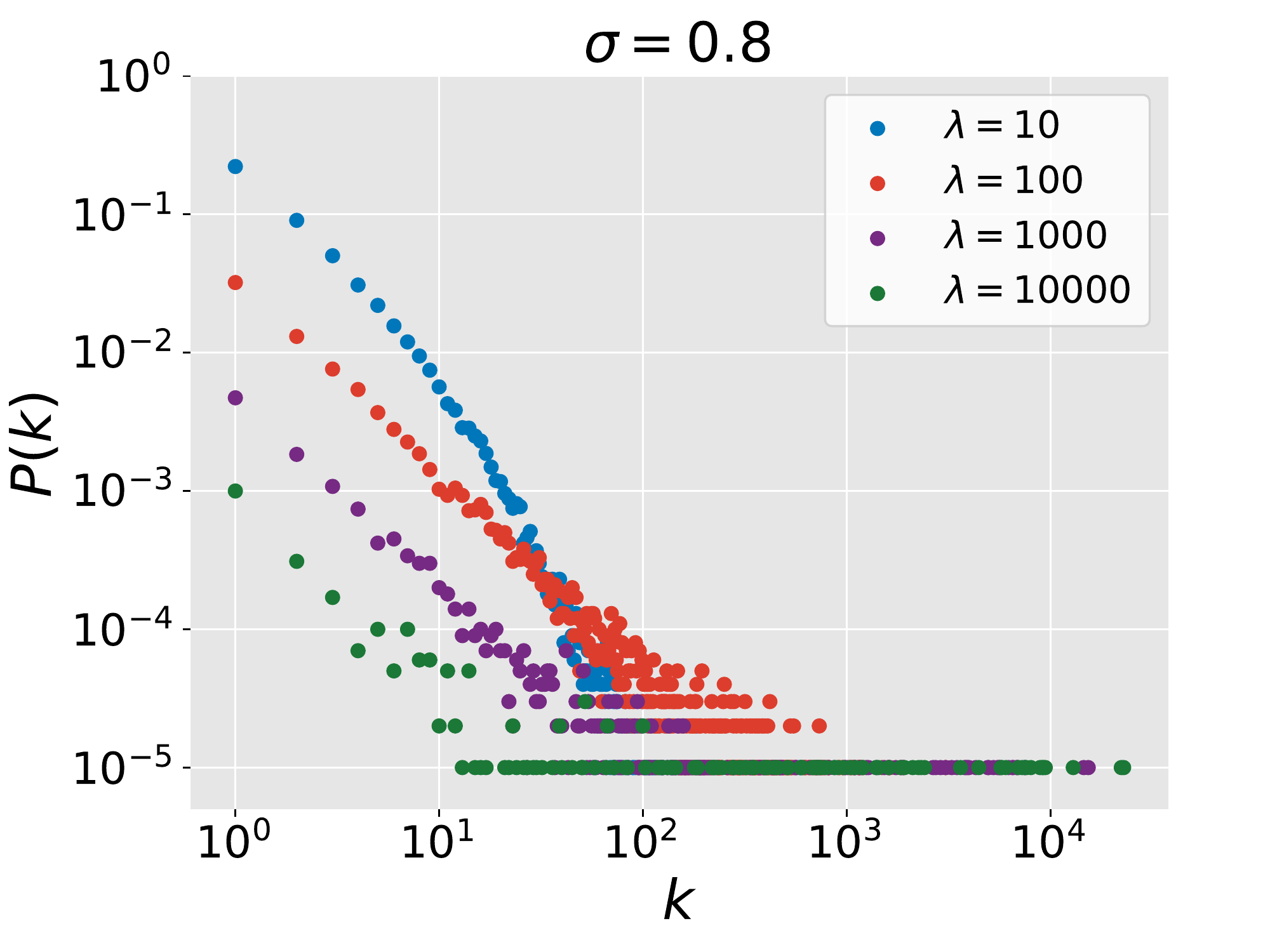}
\caption{\small{Degree distribution of simulated networks generated by \alg with $100K$ edges, for various values of $\sigma$. Plots are shown on a log-log scale. Different colors correspond to different window lengths for both decay functions, $f_1$ and $f_2$.}}\label{fig:powerlaw}
\end{figure*}

In Figure (\ref{fig:powerlaw}) we look at the degree distribution of simulated graphs generated according to \alg with a window decay function, with various settings of $\sigma$ and the window length $\lambda$, plotted on a log-log scale. As $\sigma$ increases, we see a clear power law distribution.  The shorter the window, the harder it is for the tables associated with vertices to persist long enough to obtain a high degree. Interestingly, for small values of $\lambda$, we seem to see a double power law distribution, something that has been observed in real-world communication networks. Exploring this power law behavior in more detail is an area for future research.



\section{Inference}\label{sec:inference_app}

In order to define our MCMC algorithm, we begin by introducing some auxiliary variables. Conditioned on the cluster assignments $z_i$, we follow a Chinese restaurant franchise-style representation, where we imagine each cluster as a restaurant, with two sets of tables: one for senders $s_i$ and one for recipients $r_i$. We let $g_i^s$ represent the table allocation of $s_i$, and $g_i^r$ represent the table allocation of $r_i$. All customers at a table are assigned to the same vertex; let $\eta_v$ be the total number of tables associated with vertex $v$.

Following \cite{BleiFrazier2011}, rather than sampling the cluster indicators $z_i$ directly, we instead assume each edge ``follows'' another edge, specified by a variable $c_i$.
Cluster assignments are then obtained as $z_i = i$ if $c_i=i$, or $z_i = z_{c_i}$ otherwise. 
We sample the $c_i$, $g_i^s$, and $g_i^r$, along with a representation of $H$, by iterating through the following steps:

\paragraph{Sampling $\pmb{H}$.} Inspired by augmented representation schemes for the hierarchical Dirichlet process \cite{Teh:Jordan:Beal:Blei:2006}, we represent the infinite measure $H$ using a finite-dimensional vector $(h_1, h_2, \dots, h_{V_n}, h_{+})$, where $h_v$ for $v\leq V_{n}$ is the probability mass associated with vertex $v$, and $h_{+}$ is the probability mass associated with previously unseen vertices. Note that, in practice, $V_n$ is constant during training. Following Corollary 20 of \cite{Pitman1996}, we can sample
\begin{equation*}
(h_1, \dots, h_V, h_+) \sim \mbox{Dirichlet} (\eta_1 - \sigma, \dots, \eta_V - \sigma, \gamma + V \sigma).\
\end{equation*}

\paragraph{Sampling $\pmb{g_i^s}$ and $\pmb{g_i^r}$.} Conditioned on the clusters, all allowable table assignments (\textit{i.e.} all assignments where each table is associated with a single node) have equal likelihood, so we can Gibbs sample $g_i^s$ (and similarly, $g_i^r$) based on the conditional distribution
\begin{equation}
P(g_i^s|\{z_i, s_i\}) \propto \begin{cases} \sum_{j<i:z_j=z_i, s_j=s_i} f_2(d_{ij}) & \{j<i:z_j=z_i, s_j=s_i\} \neq \emptyset \\ 
\tau h_{s_i} & \mbox{otherwise}\end{cases}.\
\label{eqn:p_g}\end{equation}
We only sample the table assignments as auxiliary variables to allow us to sample the $h_v$; we discard them before sampling the cluster assignments $z_i$.

\paragraph{Sampling $\pmb{c_i}$/$\pmb{z_i}$.} 
Following \cite{BleiFrazier2011}, we first set the $i$th link to follow itself, \textit{i.e.} $c_i=i$, and then sample a new value for $c_i$ based on the conditional probability that $c_i = j$,

\begin{equation}
P(c_i = j | E, c_{\neg i}, (h_1,\dots, h_V, h_+)) \propto
\begin{cases} f(d_{ij}) P(\mathcal{E}|c_i=j, c_{\neg i}, (h_1,\dots, h_V, h_+)) & i\neq j \\ 
\alpha P(\mathcal{E}|c_i=j, c_{\neg i}, (h_1,\dots, h_V, h_+)) & i=j\end{cases}\label{eqn:p_c}
\end{equation}
where $\mathcal{E} = \left((s_1, r_1), (s_2, r_2), \dots \right)$ represents the edge structure of the graph. We can write
\begin{equation}
P(\mathcal{E}|\{c_i\}, (h_1,\dots, h_V, h_+)) = \prod_k P_k
\end{equation}
where $P_k$ is the likelihood associated with the $k$\textsuperscript{th} cluster,

$$P_k = \prod_{i:z_i = k} P(s_i|\{s_j, r_j: j<i, z_j=k\})P(r_i|\{s_j, r_j: j<i, z_j=k\})$$

where

$$P(s_i = v|\{c_{<i}, s_{<i}, r_{<i}) \propto \tau h_v +  \sum_{j<i:s_j=v} f_2(d_{ij}).$$

In practice, we only need to calculate the likelihoods associated with clusters whose occupancy changes based on the proposed move.

Pseudo-code is given in Algorithm \ref{alg:inference}. We assume gamma priors on $\alpha$, $\tau$, $\gamma$ and both decay windows $\lambda$, and a beta prior on $\sigma$, and sample from their posterior distributions using Metropolis-Hastings, as shown in Algorithm~\ref{alg:inference}.




\begin{algorithm}[t!]
	\begin{algorithmic}[1]
	\STATE Initialize $h_v \leftarrow \frac{1}{|V|+1}$ for $v=1,\ldots,|V|$
	\STATE Initialize $z_i \leftarrow 0$, for every edge $i$
	\STATE Initialize $c_0 \leftarrow 0; c_i \leftarrow c_{i-1}$
	\FOR {each epoch}
	\FOR {$i \in \{1,\ldots,N\}$}
	\STATE Sample table allocations $g_i^s, g_i^r$ following Equation~\ref{eqn:p_g}
	\ENDFOR
	\STATE Sample $(h_1,\ldots,h_V, h_+) \sim DP(\eta_1-\sigma,\ldots, \eta_V-\sigma, \gamma+|V|\sigma )$
	\FOR {$i \in \{1,\ldots,N\}$}
	\STATE Let $c_i\leftarrow i$, and update all affected cluster assignments.
	\STATE Sample $c_i$ according to Equation~\ref{eqn:p_c}
	\STATE Update all affected cluster assignments.
	\ENDFOR
	\STATE Propose $\alpha^{*} \sim \mathcal{N}(\alpha,\nu^2)$ for some step size $\nu$
	\STATE Set $\alpha \leftarrow \alpha^*$ with probability $\min(1, \frac{\mbox{Gamma}(\alpha^*|\alpha_0, \alpha_1) \prod_i p(c_i|\alpha^*, c_{<i})}{\mbox{Gamma}(\alpha|\alpha_0, \alpha_1) \prod_i p(c_i|\alpha, c_{<i})})$
	\STATE Propose $\tau^{*} \sim \mathcal{N}(\tau,\nu^2)$
	\STATE Set $\tau \leftarrow \tau^*$ with probability $\min(1, \frac{\mbox{Gamma}(\tau^*|\tau_0, \tau_1) \prod_i p(s_i, r_i|\tau^*, \{h_v\}, s_{<i}, r_{<i}, c_{<i})}{\mbox{Gamma}(\tau|\tau_0, \tau_1) \prod_i p(s_i, r_i|\alpha, \{h_v\}, s_{<i}, r_{<i}, c_{<i})})$
	\STATE Propose $\gamma^{*} \sim \mathcal{N}(\gamma,\nu^2)$
	\STATE Set $\gamma \leftarrow \gamma^*$ with probability $\min(1, \frac{\mbox{Gamma}(\gamma^*|\gamma_0, \gamma_1) \mbox{Dirichlet}(h_1, \dots, h_V, h_+|\eta_1-\sigma, \dots, \eta_V-\sigma, \gamma^*+|V|\sigma)}{\mbox{Gamma}(\gamma|\gamma_0, \gamma_1) \mbox{Dirichlet}(h_1, \dots, h_V, h_+|\eta_1-\sigma, \dots, \eta_V-\sigma, \gamma+|V|\sigma)}$
	\STATE Propose $\sigma^{*} \sim \mathcal{N}(\gamma,\nu^2)$
	\STATE Set $\sigma \leftarrow \sigma^*$ with probability $\min(1, \frac{\mbox{Beta}(\sigma^*|\sigma_0, \sigma_1) \mbox{Dirichlet}(h_1, \dots, h_V, h_+|\eta_1-\sigma^*, \dots, \eta_V-\sigma^*, \gamma+|V|\sigma^*)}{\mbox{Beta}(\sigma|\sigma_0, \sigma_1) \mbox{Dirichlet}(h_1, \dots, h_V, h_+|\eta_1-\sigma, \dots, \eta_V-\sigma, \gamma+|V|\sigma)}$
	\STATE Propose $\lambda^{*} \sim \mathcal{N}(\lambda_{f_1},\nu^2)$ for some step size $\nu$
	\STATE Set $\lambda_{f_1} \leftarrow \lambda^*$ with probability $\min(1, \frac{\mbox{Gamma}(\lambda^*|\lambda_0, \lambda_1) \prod_i p(c_i|\lambda^*, c_{<i})}{\mbox{Gamma}(\lambda_{f_1}|\lambda_0, \lambda_1) \prod_i p(c_i|\lambda, c_{<i})})$
	\STATE Propose $\lambda^{*} \sim \mathcal{N}(\lambda_{f_2},\nu^2)$ for some step size $\nu$
	\STATE Set $\lambda_{f_2} \leftarrow \lambda^*$ with probability $\min(1, \frac{\mbox{Gamma}(\lambda^*|\lambda_0, \lambda_1) \prod_i p(s_i, r_i|\lambda^*, \{h_v\}, s_{<i}, r_{<i}, c_{<i})}{\mbox{Gamma}(\lambda_{f_2}|\lambda_0, \lambda_1) \prod_i p(s_i, r_i|\alpha, \{h_v\}, s_{<i}, r_{<i}, c_{<i})})$
	\ENDFOR
	
	\end{algorithmic}
	\caption{\alg-Inference}
	\label{alg:inference}
\end{algorithm}

\subsection{Left-to-Right Evaluation Algorithm}\label{subsec:letfright}
\begin{algorithm}[t!]
	\begin{algorithmic}[1]
		\STATE initialize $l \leftarrow 0$
		\FOR {$i=N+1, \ldots,N+n$}
		\STATE initialize $p_i \leftarrow 0$
		\FOR {$m=1\ldots M$}
		\FOR {$i' < i$}
		\STATE Sample $c_{i'}^{(m)} \sim P(c_{i'}^{(m)} |(s_{i'},r_{i'}),c_{<i'}^{(m)}, \mathcal{C} , \mathcal{C'}_{<i'},\mathcal{G},\mathcal{G}'_{<i'})$ 
		\STATE Sample $g_{i'}^{s,(m)} \sim P(g_{i'}^{s,(m)} |c_{i'}^{(m)},(s_{i'},r_{i'}),g_{<i'}^{s,(m)}, \mathcal{C} , \mathcal{C'}_{<i'},\mathcal{G},\mathcal{G}'_{<i'})$ 
		\STATE Sample $g_{i'}^{r,(m)} \sim P(g_{i'}^{r,(m)} |c_{i'}^{(m)},(s_{i'},r_{i'}),g_{<i'}^{r,(m)}, \mathcal{C} , \mathcal{C'}_{<i'},\mathcal{G},\mathcal{G}'_{<i'})$ 
		\ENDFOR
		\STATE $\begin{aligned}
              p_i \leftarrow p_i &+ \sum_k^{K_i} P(s_i,r_i|c_i^{(m)}=k, g_{i}^{s,(m)}, g_{i}^{r,(m)}) \\
                &\times P(c_i^{(m)}=k,g_{i}^{s,(m)}, g_{i}^{r,(m)} |\mathcal{C}'_{<i}, \mathcal{G}'_{<i}, \mathcal{C}, \mathcal{G})\end{aligned}$
		\ENDFOR
		\STATE $p_n \leftarrow p_n/M$
		\STATE $l \leftarrow l+\log p_n$
		\ENDFOR
	    \STATE $\log P(\mathcal{E}'|\mathcal{C},\mathcal{G}) \simeq l$
	\end{algorithmic}
	\caption{``left-to-right'' evaluation algorithm}
	\label{alg:left-to-right}
\end{algorithm}
In our experimental analysis, we compare different edge-based models in terms of their estimated test set probabilities: higher probabilities suggests we have better captured the structure underlying the data. In unsupervised models for sequences of observations---such as the graph models considered in this paper, or language models---it is hard to estimate the probability of a new sequence, since the probability of any specific sequence is vanishingly small, and often depends on the setting of latent variables such as the cluster allocations in our model, or the topic allocations in topic models. \citet{Wallach:2009:EMT:1553374.1553515} consider a number of methods to estimate these probabilities. The paper looks at the setting of topic models, but the findings generalize to sequences of edges. We use their ``left-to-right'' algorithm, which they found to be a good estimator of the true test set probability. We summarize this algorithm below, in the context of our model.

Given a sequence of edges $\mathcal{E} = (s_1,r_1),  \dots, (s_N, r_N)$, with their associated cluster assignments $\mathcal{C} = c_1, \dots, c_N$ and table assignments $\mathcal{G} = (g_1^s, g_1^r), \dots, (g_N^s, g_N^r)$, then the probability of a held-out data set $\mathcal{E}' = (s_{N+1},r_{N+1}), \dots, (s_{N+n}, r_{N+n})$ is given by

\begin{equation}\label{eqn:lda}
\begin{aligned}
    P(\mathcal{E}'|\mathcal{C}, \mathcal{G}) =& \sum_{\mathcal{C'},\mathcal{G}'} P(\mathcal{E}'|\mathcal{C'},\mathcal{G}')P(\mathcal{C'},\mathcal{G}'|\mathcal{C},\mathcal{G})  \\
    =& \prod_{i=N+1}^{N+n}\sum_{\mathcal{C}'_{<i},\mathcal{G}'_{<i}}P(s_i, r_i|c_i, g_i^s, g_i^r)P(c_i, g_i^s, g_i^r|\mathcal{C}'_{<i},\mathcal{G}'_{<i}, \mathcal{C}, \mathcal{G})
    \end{aligned}
\end{equation}
where $\mathcal{C}' = c_{N+1}, \dots, c_{N+n}$, $\mathcal{C}'_{<i} = c_{N+1}, \dots, c_{i}$, and $\mathcal{G}', \mathcal{G}'_{<i}$ are defined analogously, and the summations are over all possible values of $\mathcal{C}'$ and $\mathcal{G}'$ (or $\mathcal{C}'_{<i}$. and $\mathcal{G}'_{<i}$).

We can then approximate the terms in Equation~\ref{eqn:lda} using the sequential algorithm described in Algorithm \ref{alg:left-to-right}, where $M$ can be seen as analogous to the number of particles in a particle filtering algorithm.
\section{Additional experimental results}\label{app:results}
Table~(\ref{tbl:hitall}) expands Tables~(\ref{tbl:hits}) and (\ref{tbl:aps}) to include standard deviation across 10 repeats and results for each of the three decay functions.

\begin{table*}[ht]
\parbox{\textwidth}{
\caption{\small{hits@$\mathtt{k}$ and AP@$\mathtt{k}$ of three decay functions}\vspace{-2mm}}
\centering
\scriptsize
\resizebox{1\textwidth}{!}{
\begin{tabular}{|c|ccc|ccc|ccc|}
	\toprule
	& \multicolumn{3}{|c|}{CollegeMsg} &\multicolumn{3}{|c|}{EmailEu} & \multicolumn{3}{|c|}{SocialEv} \\ 
	\midrule
	\textit{hits@$\mathtt{k}$} & @10 & @20 & @50& @10 & @20 & @50 & @10 & @20 & @50  \\ 
	\midrule
	\alg-\textsc{MDND}& 
	0.5 $\pm$ 0.08 & 0.63 $\pm$ 0.1 & 0.96 $\pm$ 0.02 & 
	0.94 $\pm$ 0.04 & \textbf{0.96 $\pm$ 0.06} & 0.96 $\pm$ 0.05 &
    0.74 $\pm$ 0.25 &0.91 $\pm$ 0.12 & \textbf{1.0 $\pm$ 0.01}  \\
	\alg-\textsc{Window} & 
	0.32 $\pm$ 0.12 & 0.67 $\pm$ 0.18 & 0.91 $\pm$ 0.07 & 
	\textbf{0.98 $\pm$ 0.04} & 0.94 $\pm$ 0.06 & 0.97 $\pm$ 0.04 &
	0.72 $\pm$ 0.29 &0.92 $\pm$ 0.13 & 0.98 $\pm$ 0.05 \\
	\alg-\textsc{Logistic}& 
	\textbf{0.62 $\pm$ 0.17} & \textbf{0.78 $\pm$ 0.08} & \textbf{0.98 $\pm$ 0.01} & 
	0.92 $\pm$ 0.11 & 0.89 $\pm$ 0.05 & 0.94 $\pm$ 0.11&
	0.78 $\pm$ 0.22 & \textbf{0.98 $\pm$ 0.03} & \textbf{1.0 $\pm$ 0.0} \\
	\alg-\textsc{Exponential}& 
	0.55 $\pm$ 0.15 & 0.7 $\pm$ 0.07 & 0.92 $\pm$ 0.05 & 
	0.96 $\pm$ 0.08 & 0.92 $\pm$ 0.06 & \textbf{0.99 $\pm$ 0.02} &
	0.68 $\pm$ 0.28 &0.93 $\pm$ 0.13 & 0.99 $\pm$ 0.01  \\
	\midrule
	\alg-\textsc{DRGPM}& 
	0.06 $\pm$ 0.09 & 0.06 $\pm$ 0.07 & 0.06 $\pm$ 0.05& 
	0.3 $\pm$ 0.01 & 0.21 $\pm$ 0.01 & 0.23 $\pm$ 0.03&
	0.21 $\pm$ 0.4 &0.21 $\pm$ 0.4 &0.21 $\pm$ 0.4\\
	\alg-\textsc{DPGM}& 
	0.14 $\pm$ 0.07 & 0.15 $\pm$ 0.06 & 0.14 $\pm$ 0.06 & 
	0.28 $\pm$ 0.03 & 0.21 $\pm$ 0.01 & 0.25 $\pm$ 0.02 &
	0.16 $\pm$ 0.22 &0.16 $\pm$ 0.22 &0.1 $\pm$ 0.22 \\
	\alg-\textsc{DGPPF}& 
	0.17 $\pm$ 0.12 & 0.17 $\pm$ 0.09 & 0.16 $\pm$ 0.05 & 
	0.32 $\pm$ 0.01 & 0.2 $\pm$ 0.02 & 0.2 $\pm$ 0.01  &
	0.36 $\pm$ 0.47 &0.36 $\pm$ 0.47 &0.35 $\pm$ 0.47\\
	\midrule
	\textit{AP@$\mathtt{k}$} & @10 & @20 & @50& @10 & @20 & @50 & @10 & @20 & @50  \\ 
	\midrule
	\alg-\textsc{MDND}& 
	0.43 $\pm$ 0.02 & 0.33 $\pm$ 0.04 & 0.2 $\pm$ 0.03 & 
	0.66 $\pm$ 0.08 & 0.41 $\pm$ 0.03 & \textbf{0.18 $\pm$ 0.02} &
    0.17 $\pm$ 0.05 & 0.1  $\pm$ 0.03 & 0.04 $\pm$ 0.01 \\
	\alg-\textsc{Window} & 
	0.63 $\pm$ 0.07 & 0.39 $\pm$ 0.06 & 0.23 $\pm$ 0.04 & 
	0.66 $\pm$ 0.06 & 0.42 $\pm$ 0.03 & \textbf{0.18 $\pm$ 0.01} &
	\textbf{0.25 $\pm$ 0.05} & \textbf{0.16 $\pm$ 0.03} & \textbf{0.13 $\pm$ 0.01} \\
	\alg-\textsc{Logistic}& 
	0.63 $\pm$ 0.09 & \textbf{0.43 $\pm$ 0.06} &  \textbf{0.26 $\pm$ 0.04} & 
	0.66 $\pm$ 0.06 & 0.42 $\pm$ 0.06 & \textbf{0.18 $\pm$ 0.02} &
	0.18 $\pm$ 0.02 & 0.12 $\pm$ 0.03 & 0.1  $\pm$ 0.01 \\
	\alg-\textsc{Exponential}& 
	\textbf{0.64 $\pm$ 0.15} & 0.39 $\pm$ 0.09 & 0.21 $\pm$ 0.04 &
	\textbf{0.73 $\pm$ 0.06} & \textbf{0.43 $\pm$ 0.04} & \textbf{0.18 $\pm$ 0.02} &
	0.18 $\pm$ 0.03 & 0.13 $\pm$ 0.03 & 0.09 $\pm$ 0.02  \\
	\midrule
	\alg-\textsc{DRGPM}& 
	0.09 $\pm$ 0.03 & 0.05 $\pm$ 0.01 & 0.02 $\pm$ 0.01& 
	0.21 $\pm$ 0.0 & 0.14 $\pm$ 0.0 & 0.1 $\pm$ 0.0  &
	0.06 $\pm$ 0.05 & 0.03 $\pm$ 0.02 & 0.01 $\pm$ 0.01\\
	\alg-\textsc{DPGM}& 
	0.1 $\pm$ 0.02 & 0.05 $\pm$ 0.01 & 0.02 $\pm$ 0.0& 
	0.16 $\pm$ 0.0 & 0.14 $\pm$ 0.0 & 0.12 $\pm$ 0.0 &
	0.05 $\pm$ 0.06 & 0.02 $\pm$ 0.03 & 0.01 $\pm$ 0.01\\
	\alg-\textsc{DGPPF}& 
	0.1 $\pm$ 0.0 & 0.05 $\pm$ 0.0 & 0.02 $\pm$ 0.0& 
	0.22 $\pm$ 0.0 & 0.17 $\pm$ 0.0 & 0.1 $\pm$ 0.0&
	0.06 $\pm$ 0.05 & 0.03 $\pm$ 0.03 & 0.01 $\pm$ 0.01\\
	\hline
\end{tabular}
}
\label{tbl:hitall}
}
\end{table*}

\end{document}